\documentclass[10pt,twocolumn,letterpaper]{article}

\usepackage[pagenumbers]{cvpr} 
\usepackage{enumitem}

\definecolor{cvprblue}{rgb}{0.21,0.49,0.74}
\usepackage[pagebackref,breaklinks,colorlinks,citecolor=cvprblue]{hyperref}
\usepackage{caption}
\usepackage{colortbl}
\usepackage{booktabs}
\usepackage{marvosym}
\usepackage{lipsum}

\def\paperID{8660} 
\def\confName{CVPR}
\def\confYear{2025}
\def\oursEnhancer{InstanceEnhancer}
\newcommand{\oursName}{$\mathtt{InstanceCap}$}
\newcommand{\oursData}{$\mathtt{InstanceVid}$}

\title{InstanceCap: Improving Text-to-Video Generation via Instance-aware Structured Caption}

\author{%
    \textbf{Tiehan Fan}$^{1}$\footnotemark[1]\quad
    \textbf{Kepan Nan}$^{1}$\footnotemark[1]\quad
    \textbf{Rui Xie}$^{1}$\quad
    \textbf{Penghao Zhou}$^{2}$\quad \\
    \textbf{Zhenheng Yang}$^{2}$\quad
    \textbf{Chaoyou Fu}$^{1}$\quad
    \textbf{Xiang Li}$^{3}$\quad
    \textbf{Jian Yang}$^{1}$\quad
    \textbf{Ying Tai}$^{1}$\textsuperscript{\Letter} \\
    $^1$ Nanjing University\quad
    $^2$ ByteDance \quad
    $^3$ Nankai University\\
    \url{https://github.com/NJU-PCALab/InstanceCap}
}

\begin{document}
\maketitle
\newcommand\blfootnote[1]{%
\begingroup
\renewcommand\thefootnote{}\footnote{#1}%
\addtocounter{footnote}{-1}%
\endgroup
}

\blfootnote{*~Equal contributions.~Ying Tai is the corresponding author.}

\begin{abstract}
Text-to-video generation has evolved rapidly in recent years, delivering remarkable results.
Training typically relies on video-caption paired data, which plays a crucial role in enhancing generation performance. However, current video captions often suffer from insufficient details, hallucinations and imprecise motion depiction, affecting the fidelity and consistency of generated videos. In this work, we propose a novel instance-aware structured caption framework, termed~\oursName, to achieve instance-level and fine-grained video caption for the first time.
Based on this scheme, we design an auxiliary models cluster to convert original video into instances to enhance instance fidelity. Video instances are further used to refine dense prompts into structured phrases, achieving concise yet precise descriptions. Furthermore, a $22$K~\oursData~dataset is curated for training, and an enhancement pipeline that tailored to~\oursName~structure is proposed for inference. Experimental results demonstrate that our proposed~\oursName~significantly outperform previous models, ensuring high fidelity between captions and videos while reducing hallucinations.

\end{abstract}
    
\section{Introduction}
\label{sec:intro}
Recently, text-to-video (T2V) generation with advanced diffusion transformers (DiT) \cite{open-sora-plan, zeng2024open, ma2024latte, chen2024videocrafter2, li2024t2v, jin2024pyramidal, kuaishou2024kling, pika2023pika, runway2023gen, wang2023modelscope, zhang2023show, wang2023lavie} have attracted significant attention for the ability to generate realistic, long-duration videos based on text prompts.
Video-caption paired data is typically used in training and plays a crucial role in enhancing generation performance.
Current video recaption methods often incorporate multimodal large language models to produce detailed captions, which however usually suffer from hallucinations, leading to inconsistencies between captions and video content.
Consequently, creating \textit{consistent video-caption pairs with accurate details and precise motion depiction} for T2V generation remains a significant challenge.

\begin{figure*}[!t]
	\centering
	\includegraphics[width=\linewidth]{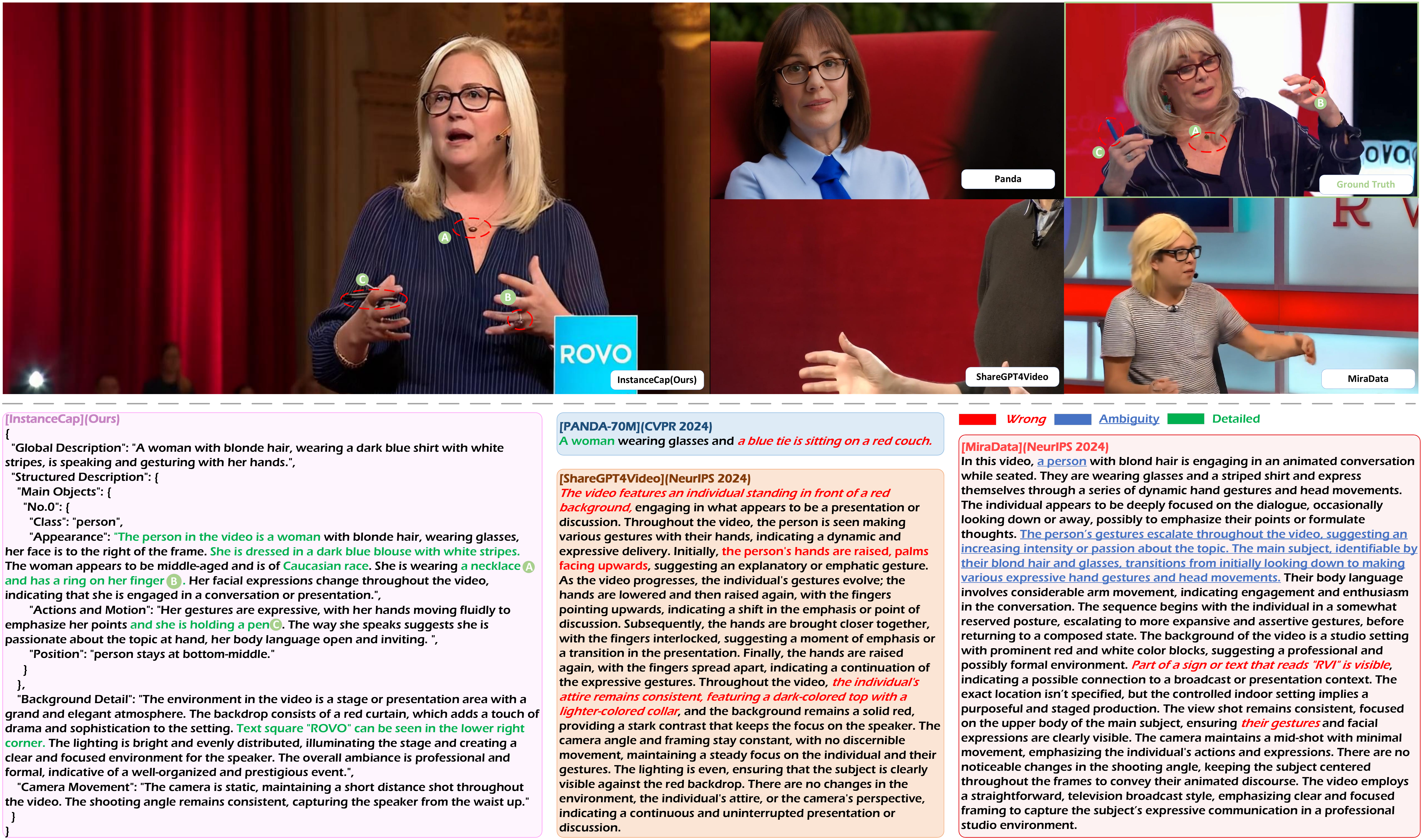}
	\vspace{-6mm}
	\caption{\textbf{Top:} Comparison of the reconstruction-via-recaption results between~\oursName~and state-of-the-art captioning methods for annotating the ground truth video.~\oursName~produces results that more closely resemble the original video, showing greater detail fidelity (highlighted by the red circle). \textbf{Bottom:} The corresponding captions generated by~\oursName~and others. \textit{\textcolor{red}{Red}} denotes incorrect captions, \underline{\textcolor{blue}{blue}} represents ambiguous captions, and \textcolor{green}{green} indicates detailed and accurate descriptions of video. Specific visual hints are marked as A, B, and C for clarity.
    All videos are generated using the \textit{same} video generation product, \textbf{Hailuo AI}\protect\footnotemark, which has robust prompt-following capabilities, clearly highlighting the effectiveness of~\oursName.
    }
	\label{fig:compare1}
	\vspace{-2mm}
\end{figure*}

As shown in Figure~\ref{fig:compare1}, current video recaption methods can be broadly categorized into three types:
$1$) Short captions, such as Panda-70M \cite{chen2024panda}, lack sufficient coverage of video content, leading to low fidelity.
$2$) Dense captions, like ShareGPT4Video \cite{chen2024sharegpt4video}, enrich textual content but suffer from hallucination issues, often generating meaningless or inaccurate video content.
$3$) Coarse-level structured captions, exemplified by MiraData \cite{ju2024miradata}, improve video quality but provide coarse-level details. Moreover, the redundancy introduced by MLLM across structures diminishes its overall effectiveness.
To this end, achieving accurate captions remains two crucial challenges:
1) \textit{High fidelity between caption and video}: Retain as much of the original video’s objects, textures, and motion information as possible.
2) \textit{Accurate content in caption}: Enable MLLM model to generate precise content, minimizing hallucinations and repetition.

To address the challenges, we propose a novel instance-aware structured caption framework, termed~\oursName, to achieve instance-level and fine-grained video caption for the first time.
Our structure is specifically designed to incorporate \textit{instances, background, and camera movement}. 
For each instance, we specify \textit{class, appearance, actions, motion, and position}.
To enhance the fidelity and accuracy of video captions, we focus on two key aspects:
1) \textit{From Global Video to Local Instances}:
For each instance, we propose an 
auxiliary models cluster (AMC) to isolate it from the original video and obtain the corresponding position and category information. This operation minimizes interference from unrelated regions while retaining as much of the original video's information as possible.
2) \textit{From Dense prompt to Structured Phrases}:
We leverage multimodal large language models (MLLMs) in an improved Chain-of-Thought (CoT) process to obtain concise yet accurate descriptions of textures, camera movement, actions and motion for each instance.
This reduces the probability of hallucinations and irrelevant content produced by the language model compared to traditional caption methods that directly describe video content in a complex, dense caption.

To validate the effectiveness of~\oursName~in T2V generation, we constructed a high-definition video dataset comprising 22K samples to create a training dataset with our instance-aware structured captions, named~\oursData. 
At the inference stage, we also implemented a prompt enhancement pipeline tailored to our structured captioning method, enabling the generation of concise captions that better align with user needs.
Our~\oursName~integrates seamlessly with existing diffusion models. 
Experimental results demonstrate that after finetuned with our~\oursData, the T2V model exhibits better ability with prompt following on details and motion actions. 
In summary, our main contributions are as follows:

\begin{itemize}[left=0pt]
\item We propose~\oursName, the first instance-aware structured caption method for text-to-video generation. 
We also constructed a $22$K~\oursData~dataset during training, and developed an enhancement pipeline that tailored to~\oursName~structure during inference.

\item 
%
We design the AMC paradigm to convert global video into instances, enhancing instance fidelity. Additionally, we propose an improved CoT pipeline with MLLMs to refine dense prompts into structured phrases, achieving concise yet precise descriptions.

\item Extensive experiments on video reconstruction demonstrate that our \oursName~significantly enhances the fidelity between captions and videos. T2V models fine-tuned on our~\oursData~further achieve more precise generation on instance details and motion actions.

\footnotetext{https://hailuoai.com/video}
\end{itemize}

\section{Related works}
\label{sec:related_works}

\paragraph{Video recaptioning.}
Advancements in text-to-video generation demand high-quality video-text datasets to build robust foundational video models for visual-language alignment. 
Current video recaption methods fall into two main categories: manual annotation~\cite{zhou2018YOUCOOK, chen:acl11, wang2020vatex} and end-to-end recaption using multimodal large language models (MLLM). 
Although manual annotation provides higher accuracy, scaling datasets to meet the needs of high-quality video generation models remains a substantial challenge.
Recent advances in MLLM have demonstrated impressive capabilities in video understanding and description generation.
Panda~\cite{chen2024panda} and InternVid~\cite{wang2023internvid} are with short captions, offering computational efficiency but frequently omitting crucial content and exhibit low video fidelity. 
OpenVid-1M~\cite{nan2024openvid}, Vript~\cite{yang2024vriptvideoworththousands} and ShareGPT4Video~\cite{chen2024sharegpt4video} are with dense captions, which provide richer content but face challenges: Hallucinations due to complexity, inclusion of redundant information, and text encoder truncation caused by excessively long text.
MiraData~\cite{ju2024miradata} achieves coarse-level structured captions that attempts to mediate these issues but struggle with fine detail and redundancy across structures.
Different from the existing video recaption methods,~\oursName~is the first \textit{instance-aware structured caption} approach for text-to-video generation,  ensuring high fidelity between caption and video while reducing hallucinations and repetition.

\begin{figure*}[t!]
	\centering
	\includegraphics[width=0.95\linewidth]{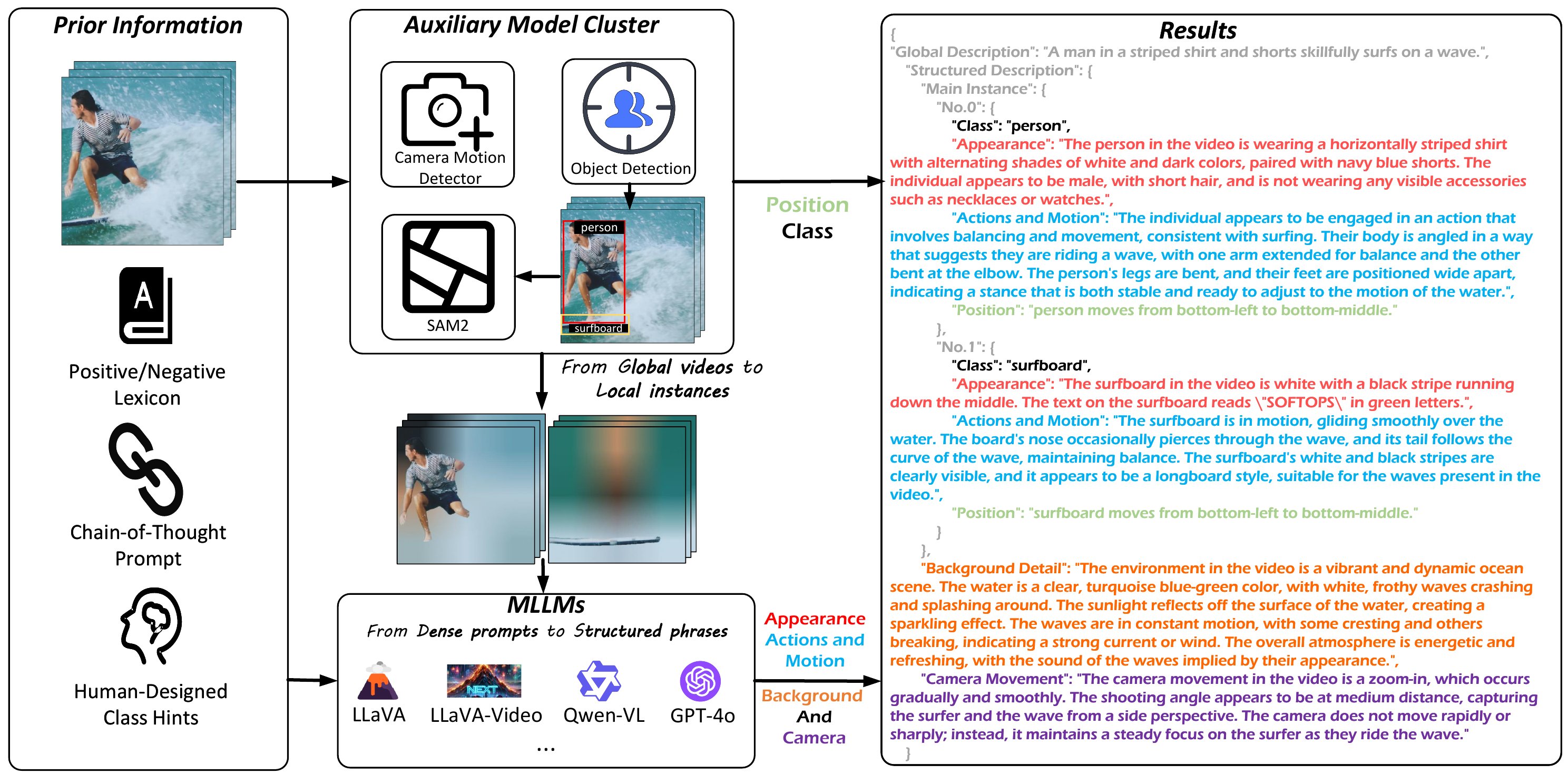}
	\vspace{-2mm}
	\caption{\textbf{Overview of InstanceCap pipeline}. Details of ``\textit{from dense prompts to structured phrases}'' design are shown in Figure~\ref{fig:mllm}.}
	\label{fig:pipeline}
	\vspace{-3mm}
\end{figure*}

\vspace{-4mm}
\paragraph{Text-to-video generation.}

Despite numerous high-quality video generation models~\cite{open-sora-plan, chen2024videocrafter2, jin2024pyramidal, kuaishou2024kling, pika2023pika, yang2024cogvideox} perform well with simple directives, they often falter with complex prompts requiring precise instance-level details or intricate camera movements.
Analyzing current video-text datasets suggests these limitations may stem from suboptimal training data quality. Traditional recaption methods have not sufficiently captured instance-specific detail granularity or provided comprehensive descriptions of camera movements. 
To enhance instance-level details and motion consistency, we construct a $22$K~\oursData~dataset for training/finetuning, and develope an enhancement pipeline~\oursEnhancer~that tailored to the proposed instance-aware structure during inference. 

\section{Method}
\label{sec:method}

In Section~\ref{InstanceCap}, we first present the~\oursName~pipeline, as shown in Figure~\ref{fig:pipeline}. 
Based on this pipeline, we recaption the carefully curated dataset~\oursData~in Section~\ref{InstanceVid dataset}, enhancing T2V models' instance generation. 
Additionally, in Section~\ref{InstanceEnhancer}, we introduce~\oursEnhancer~to convert short prompts into our proposed instance-aware structured caption format during inference.

\subsection{InstanceCap}
\label{InstanceCap}

\paragraph{Video preprocessing with auxiliary model cluster.} 
For continuous video processing, we implemented uniform sampling using decord\footnote{https://github.com/dmlc/decord}, following the methodology established in LLaVA-Video~\cite{zhang2024videoinstructiontuningsynthetic}. 
This approach enables us to extract essential temporal metadata, including duration, frame count, and timestamps, allowing MLLMs to better interpret temporal sequences in recaptioning tasks. 
Additionally, to enhance MLLMs' capabilities through structured guidance, we incorporate several auxiliary models to achieve accurate object detection, video instance segmentation and camera motion prediction, providing precise prior information to the MLLMs.

\vspace{-4.5mm}
\paragraph{Global description, background detail and camera movement.}  
\label{sec:g_caption}
When describing video content, a high-quality global description should capture primary elements, environmental context, camera movements, angles, and tonal qualities.
MLLMs excel in generating high-level video summaries using Chain-of-Thought methodology.
By employing carefully designed prompts with CoT, we can guide MLLMs to produce detailed background descriptions while minimizing references to foreground elements.

However, MLLMs’ limitations in processing discrete frames rather than continuous video segments make it challenging to \textit{distinguish camera motion from instance action}. 
To address this, we achieve camera annotations from Open-Sora~\cite{opensora} for basic movements (\textit{e.g.}, zoom, rotation) and rely on MLLMs to capture subtle motion attributes (\textit{e.g.}, intensity, speed).
The integration of camera movement indicators with MLLM capabilities provides comprehensive annotations, as illustrated in Figure~\ref{fig:camera_human} (a). 

\begin{figure*}[t!]
    \centering
    \includegraphics[width=0.84\linewidth]{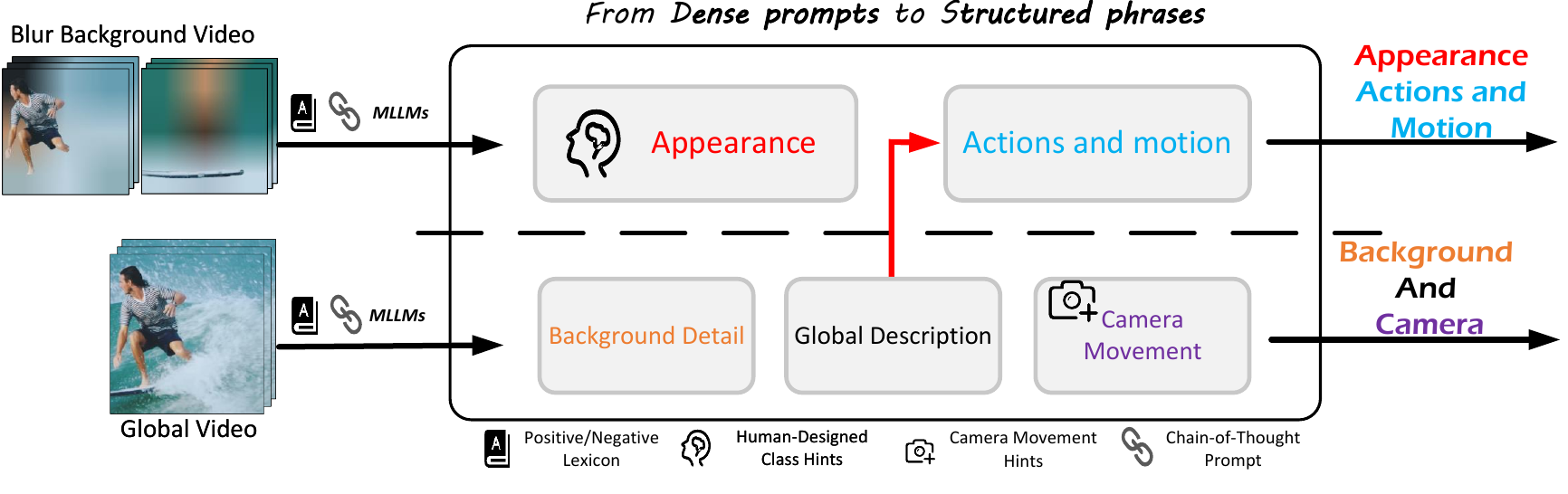}
    \vspace{-3.5mm}
    \caption{Details on ``\textit{from dense prompts to structured phrases}'' design. We propose an improved CoT pipeline with carefully designed information interactions (\textcolor{red}{red arrow}), which facilitates MLLMs to accurately capture instances with precise descriptions on attributes.}
    \label{fig:mllm}
\end{figure*}

\vspace{-4.5mm}
\paragraph{Structured Description on Instances.}



In this subsection, we introduce the details to achieve our instance-aware structured description.
To address MLLMs' limitations in instance annotation and the suboptimal results of directly adapting \textit{weak visual prompts} from images to videos~\cite{YAO202430, Shtedritski_2023_ICCV}, we make full use of the auxiliary model cluster, including initial object detection~\cite{zong2023detrs}, video instance segmentation with SAM2~\cite{ravi2024sam2}, and blur non-instance regions to achieve blur background, resulting in better recaptioning outcomes compared to alternative visual prompt methods in video, as shown  in Figure~\ref{fig:background}.
To this end, we \textit{decompose the global videos into local instances}.

Next, we describe how to achieve detailed and accurate description of each instance (Figure~\ref{fig:mllm}). To maintain instance-level precision, we deliberately constrain the information accessible to MLLMs during instance annotation. 
Crucially, the global video remains \textbf{invisible} due to our designed blurred backgrounds, 
preventing MLLMs from \textit{confusing information across multiple instances}. 
This approach allows~\oursName~to focus on local instances identified through auxiliary model cluster.
Furthermore, to avoid the potential limitation of MLLMs seeing only isolated instances, which could lead to overlooking inter-instance interactions and subsequent misinterpretations, we incorporate the global description mentioned in previous subsection. Specifically, we \textit{inject the global description into the instance-annotation MLLMs}. This strategic mitigates potential biases in instance action descriptions while maintaining instance-specific accuracy.

To enhance the capability of~\oursName~in capturing instance-level details, we introduce novel insights into the improved CoT process. Our analysis of current MLLM-based video recaption methods shows that simple prompts, like ``\textit{Please provide a detailed description of this video}'' or Chain-of-Thought prompts ``\textit{Let's think step by step... First, please note... Finally, summarize the video content...}'' fail to capture precise instance details. 
Additional experiments reveal that MLLMs can effectively annotate these details when given fine-grained prompts, such as `` \textit{Please note if the characters have any accessories}'' or ``\textit{Please observe whether there are spots on the bananas}''.
To enhance the details of instances, we develope \textbf{Human-designed Class Hints}, crafting specific prompts for about $80$ detectable categories using our auxiliary models cluster.
Specifically,  we present the ``person" class prompt here: ``\textit{Please focus primarily on the person's facial expressions, attire, age, gender, and race in the video and give a detailed description. Please mention if there are any necklaces, watches, hat or other decoration; otherwise, there's no need to bring them up}.''
Besides, we also developed a curated \textbf{Positive/Negative Lexicon} to guide MLLMs in generating more aesthetically refined captions. 
More details can be found in our supplementary material.

\subsection{InstanceVid}
\label{InstanceVid dataset}

\paragraph{Data collection.}

\oursData~is curated via refining a subset from the high-aesthetic, high-consistency videos from OpenVid-1M~\cite{nan2024openvid}. 
To showcase our method’s high-fidelity labeling of instance details and motion, we selected video samples that included at least one instance exhibiting high motion intensity during dataset filtering.

\begin{figure}[t!]
	\centering
	\includegraphics[width=1\linewidth]{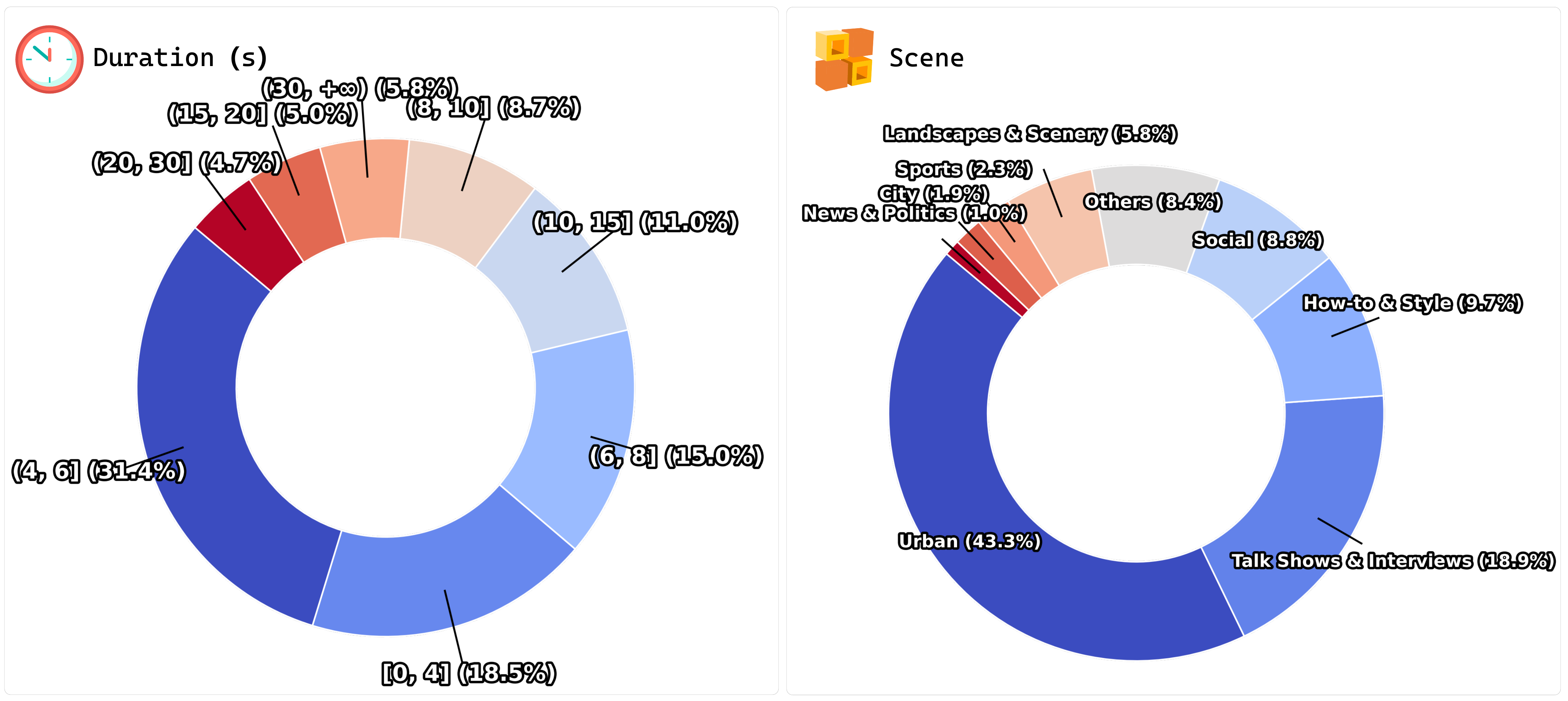}
	\vspace{-5mm}
	\caption{\oursData~provides structured captions for videos in open-domain scenarios, featuring diverse instance, expansive scenes, precise and instance-aware captions, and video-generation-friendly durations.}
	\label{fig:circle}
	\vspace{-6mm}
\end{figure}

\vspace{-3mm}
\paragraph{Statistical analysis of~\oursData.}
Figure~\ref{fig:circle} illustrates the statistical characteristics of~\oursData across two main dimensions: video scenes, and temporal durations. 
Our data collection emphasizes videos with distinct instances while ensuring a balanced representation of outdoor scenes to prevent biases from an overemphasis on instance-focused content. 
We achieve detailed descriptions capturing human movements, physical appearances, and documentation of common objects and animals.
Besides, \oursData~focuses on short-duration videos (2-10 seconds) for two main reasons. 
First, OpenVid-1M segments longer sequences to eliminate excessive scene transitions. 
Second, most of the current open-source T2V models are optimized for video generation within this duration range.

\subsection{InstanceEnhancer}
\label{InstanceEnhancer}
When the caption distribution of training data differs from that of inference text, it may result in poor instruction-following performance or even problematic outputs. 
This issue is evident in T2V generation, particularly when long captions are used for training but short captions for inference, leading to subpar results.
Since users typically prefer short captions, it is essential to enhance short caption effectively to better align with our proposed instance-aware structured caption during training.

\begin{figure}[t!]
    \centering
    \includegraphics[width=1\linewidth]{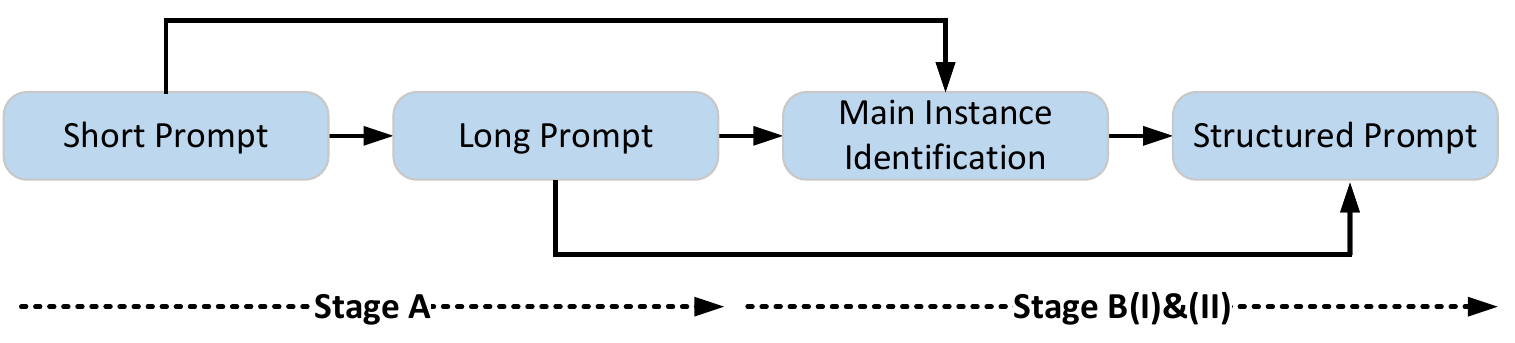}
    \vspace{-7mm}
    \caption{High-level overview of~\oursEnhancer, illustrating the data flow and the partitioning of stages. For a detailed implementation, refer to the supplemental materials, which provide an in-depth description of the enhancer pipeline design and the interdependencies between the stages.}
    \vspace{-4mm}
    \label{fig:enhancer_simple}
\end{figure}

As shown in Figure~\ref{fig:enhancer_simple}, we introduce a tuning-free approach called~\oursEnhancer~that achieves this by strictly limiting the generated formats to match the caption corresponding to the training input we used. Our method differs from existing tuning-free caption enhancement approaches, such as those presented in RPG~\cite{yang2024mastering}. Instead of directly enhancing short captions, which we found can introduce inconsistencies between multiple instances' actions and their environmental context in video generation, we employ a two-stage enhancement strategy.
In Stage A, short prompts are expanded into detailed long prompts. 
Stage B(I)\&(II) uses both expanded and original captions to segment and enhance specific instances, preserving contextual coherence while ensuring precise instance identification. 
Due to the space limitation, more details of our enhance pipeline can be found in the supplementary material.

\section{Experiments}
\label{sec:experiments}

\begin{figure}[t!]
    \centering
    \includegraphics[width=1\linewidth]{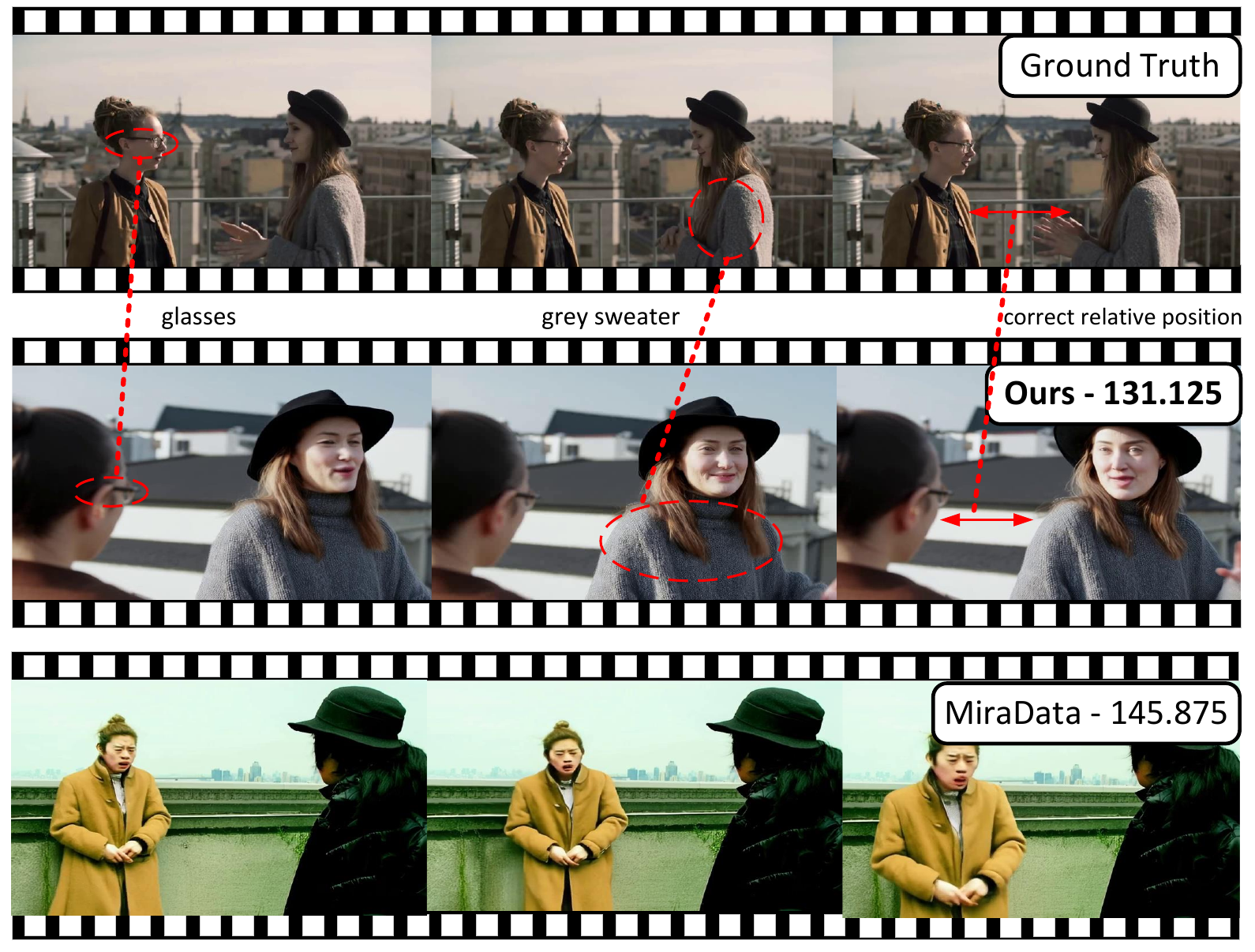}
    \vspace{-7mm}
    \caption{Comparison on reconstruction-via-recaption  between~\oursName~and MiraData. 
    Corresponding 3DVAE scores are also indicated. 
    Similar semantics shared between~\oursName~and GT are indicated by red circles and lines.}
    \label{fig:vae}
    \vspace{-2mm}
\end{figure}

\begin{figure*}[t!]
    \centering
    \includegraphics[width=0.96\linewidth]{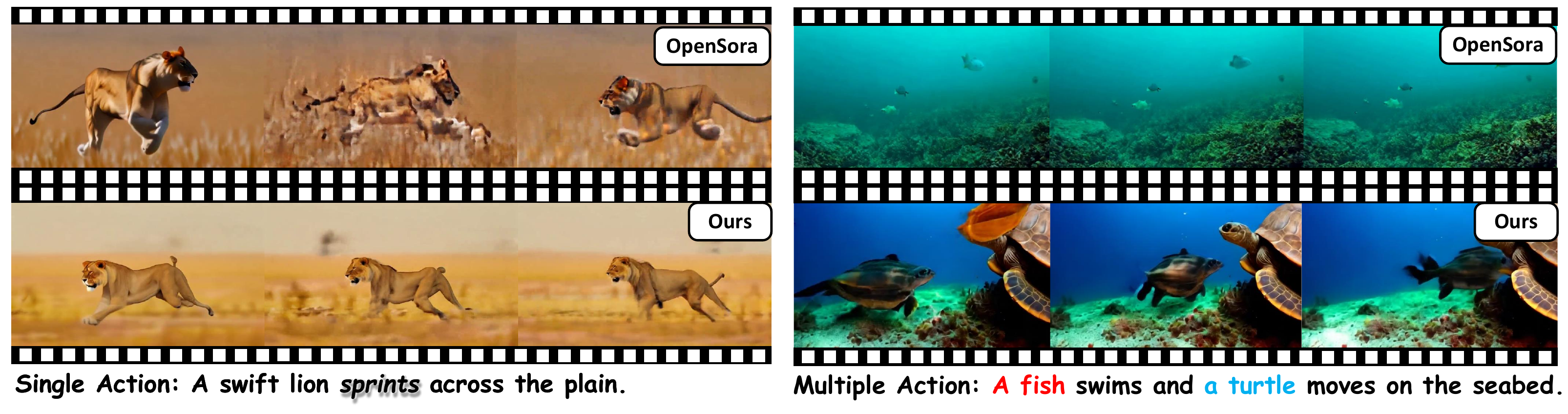}
    \vspace{-4mm}
    \caption{Visual comparison of~\oursName~and Opensora on Single and Multiple Action Score. In terms of the dynamic degree of video generation, we show better consistency and enhanced multi-instance dynamic generation effect.}
    \label{fig:action}
    \vspace{-1mm}
\end{figure*}

\subsection{Experimental setup}
\paragraph{Video reconstruction with recaptions.}
To comprehensively evaluate~\oursName, we conducted a series of experiments, benchmarking against state-of-the-art methods including Panda-70M \cite{chen2024panda}, ShareGPT4Video \cite{chen2024sharegpt4video}, and MiraData \cite{ju2024miradata}.
To this end, we carefully selected $100$ video clips from OpenVid-1M~\cite{nan2024openvid} and Animal Kingdom~\cite{Ng_2022_CVPR}. 
For each video, we generated one caption using various caption models, which were then input into the advanced T2V model CogvideoX-5b~\cite{yang2024cogvideox} for video generation. 
We calculated the differences between the generated videos and the ground truth videos to evaluate each caption model's performance, where smaller visual differences indicate more accurate captions and higher fidelity.

We introduced several metrics to evaluate the video reconstruction performance:
$1$) $\text{3DVAE}_{score}$: Using 3DVAE from CogVideoX~\cite{yang2024cogvideox} as the backbone, we extract hidden-space representations from both the original videos and their recaption-reconstructed counterparts. 
These representations quantify the perceptual distance between them.
$2$) $\text{CLIP}_{SenbySen}$: To handle CLIP’s 77-token processing limit, we segment long captions into individual sentences and compute CLIP \cite{radford2021learningtransferablevisualmodels} similarity between each sentence and every original video frame. The final score is obtained by first averaging the similarity scores of each sentence across all frames, then averaging these sentence-level scores for a comprehensive result.
$3$) Human Evaluation: We conducted a user study with a panel of evaluators to assess caption quality across two aspects: Instance Detail (ID) and Hallucination Scores (HS).

\begin{figure*}[t!]
    \centering
    \begin{minipage}{0.5\textwidth}
        \centering
        \vspace{+2mm}
        \resizebox{1\linewidth}{!}{%
        \begin{tabular}{l|cc|cc|c}
            \toprule
            Captioning Methods & $\text{3DVAE}_{score}$↓ & $\text{CLIP}_{SenbySen}$↑  & Avg. Length \\
            \midrule
            Panda-70M           & 140.25 & 0.1956  & 13 words \\
            ShareGPT4Video       & 141.00 & 0.2132  & 191 words \\
            LLaVA-Video-72B     & 139.88 & 0.2060 &102 words \\
            MiraData(GPT-4o)    & \underline{137.50} & \textbf{0.2156}  & 263 words \\
            \textbf{\oursName (Ours)} & \textbf{134.25} & \underline{0.2133} & 157 words \\
            \bottomrule
        \end{tabular}%
        }
        \vspace{+1mm}
        \captionof{table}{Quantitative comparisons on reconstruction-via-recaption results. The best results are marked in \textbf{bold}, and the second-best are \underline{underscored}. 
        As a reference, CogVideoX-5b accepts $226$ text tokens, with any excess being truncated.}
        \label{tab:reconstruction}
    \end{minipage}%
    \hfill
    \begin{minipage}{0.48\textwidth}
        \centering
        \vspace{-8mm}
        \includegraphics[height=0.39\linewidth]{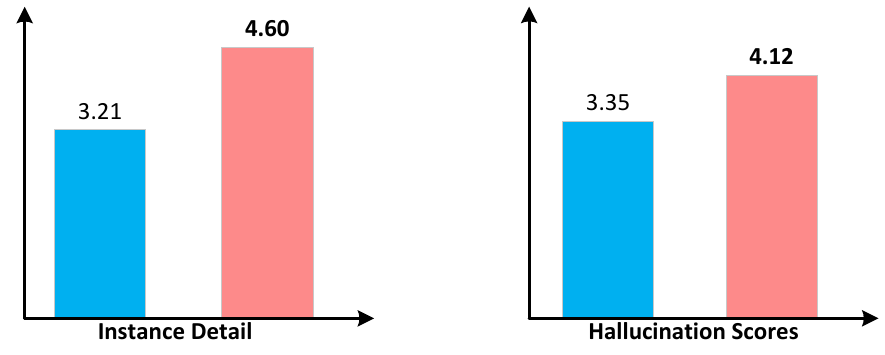}
        \vspace{-0.8mm}
        \caption{User study on instance detail and hallucination scores.
        	Our instance-aware structured caption shows clear advantages compared to the coarse-structured MiraData~\cite{ju2024miradata}. 
        	}
        \label{fig:human_eval}
    \end{minipage}
\end{figure*}

\vspace{-2mm}
\paragraph{T2V generation.}
To thoroughly evaluate the T2V generation performance of our~\oursName,
we utilize the~\oursData~dataset to finetune the state-of-the-art DiT-based T2V generation model Open-Sora~\cite{opensora}. 
In our evaluation, we compare with Open-Sora, CogVideoX-5b~\cite{yang2024cogvideox}, Pyramid-Flow \cite{jin2024pyramidal}, and Open-Sora-Plan \cite{open-sora-plan}. 
To enable fine-grained, instance-level assessment, we construct a highly challenging evaluation benchmark called Inseval, inspired by recent advancements in T2I and T2V evaluation~\cite{huang2023t2icompbench, lin2024evaluatingtexttovisualgenerationimagetotext, wu2024ifadapterinstancefeaturecontrol, liu2023evalcrafter, huang2023vbench}.
Specifically, we curate a diverse evaluation dataset of over $200$ carefully crafted instance-level prompt-answer pairs, covering both single-object and multi-object scenarios systematically across five key dimensions: Action, Color, Shape, Texture, and Detail. 
Motivated by the previous evaluation benchmarks~\cite{huang2023vbench}, we implement a CoT reasoning framework for generating structured QA responses to ensure objective and consistent evaluation, allowing us to derive instance-level evaluation scores that align closely with human perception and preferences. 
This approach provides a more nuanced and reliable assessment of instance-level generation quality.

\begin{table*}[t!]
\vspace{-2mm}
\centering
\resizebox{0.7\linewidth}{!}{%
\begin{tabular}{c|ccccc|ccc|c}
\toprule
T2V Model & \multicolumn{5}{c}{Single↑} &  \multicolumn{3}{c}{Multiple↑} & Average↑ \\
\cmidrule(lr){2-6} \cmidrule(lr){7-9} \cmidrule(lr){10-10}
 & Action & Color & Shape & Texture & Detail & Action & Color & Texture & \\
\midrule
    \rowcolor{gray!20}
    CogVideoX-5B~\cite{yang2024cogvideox} & 64\% & 60\% & 44\% & 60\% & 20\% & 8\% & 48\% & 40\% & 43.00\% \\
    \rowcolor{gray!20}
    Pyramid-Flow-2B~\cite{jin2024pyramidal} & 44\% & 68\% & 32\% & 32\% & 7\% & 4\% & 24\% & 16\% & 28.38\% \\
    \rowcolor{gray!20}
    Open-Sora Plan v1.3-2.7B~\cite{open-sora-plan} & 64\% & 44\% & 36\% & 32\%  & 27\% & 20\% & 32\% & 12\% & 33.38\% \\
    \midrule
    Open-Sora v1.2-1.1B~\cite{opensora} & 40\% & \underline{56\%} & \underline{36\%} & \underline{40\%} & 13\% & 12\% & 16\% & 16\% & 28.63\% \\
    ~+ \textbf{\oursName(Ours)} & \textbf{56\%} & \textbf{60\%} &  \textbf{40\%} &  \textbf{48\%} & \textbf{27\%} & \textbf{16\%} &  \textbf{32\%} &  \textbf{24\%} & \textbf{37.88\%} \\
    ~+ Panda-captioner~\cite{chen2024panda} & 40\% & 48\% & 28\% & 40\%  & 20\% & 8\% & 20\% & 12\% & 27.00\% \\
    ~+ ShareGPT4Video~\cite{chen2024sharegpt4video} & 40\% & 44\% & 32\% & 24\% & 13\% & \textbf{16\%} & 8\% & \underline{20\%} & 24.63\% \\
    ~+ LLaVA~\cite{nan2024openvid} & \underline{52\%} & 52\% & 28\% & 28\% & \underline{20\%} & 12\% & \underline{28\%} & 16\% & \underline{29.50\%} \\
\bottomrule
\end{tabular}%
}
\vspace{-2mm}
\caption{Quantitative comparison between~\oursName~and SOTA video captioning models, all based on the popular T2V model Open-Sora. Additionally, we also compare three powerful T2V models, including CogVideoX-5B, Pyramid-Flow, and Open-Sora Plan. The best results of video captioning methods and Open-Sora are marked in \textbf{bold}, and the second-best are \underline{underscored}.}
\label{tab:inseval}
\end{table*}

\subsection{Comparison with SOTA caption models}
\paragraph{Qualitative evaluation.} 
CogVideoX-5b \cite{yang2024cogvideox} is a latent text-to-video generation model known for its capability to generate realistic, long-duration videos based on text prompts. With the integration of our~\oursName~into CogVideoX-5b, as substantiated by Figure~\ref{fig:vae}, the model exhibits a notable enhancement in the video reconstruction capacity. This demonstrates that our instance-aware structured captions retain more of the original video’s information, leading to higher fidelity. For instance, our~\oursName~can retain information such as ``glasses", ``grey sweater", and ``relative position of two people", whereas MiraData \cite{ju2024miradata} almost completely loses these important details. A similar conclusion can be drawn from Figure~\ref{fig:compare1}. These results underscore the significant improvements achieved by our~\oursName, resulting in high-quality reconstruction characterized by rich detail and high fidelity between our captions and the original videos.

\vspace{-2mm}
\paragraph{Quantitative evaluation.} 
Table~\ref{tab:reconstruction} presents quality comparisons between our~\oursName~and other caption methods across two metrics. Based on the results, we make the following observations:
1) Our method delivers comparable or superior quality to the four baselines, demonstrating its ability to enhance fidelity between videos and captions.  This strong alignment with human perceptual judgments and preferences is evident in Figures~\ref{fig:compare1} and~\ref{fig:vae}.
2) Our method consistently excels across all metrics for captions under $200$ words, an acceptable length for most T2V models, illustrating its generalizability.
Figure~\ref{fig:human_eval} compares MiraData's captions with our instance-aware structured ones. We randomly selected videos from an open-domain dataset, and a panel of evaluators assessed caption quality using standardized criteria. 
Results indicate that our captions offer significantly richer and more accurate descriptions while reducing hallucination artifacts compared to MiraData’s output.

\begin{figure}[t!]
	\centering
	\includegraphics[width=0.92\linewidth]{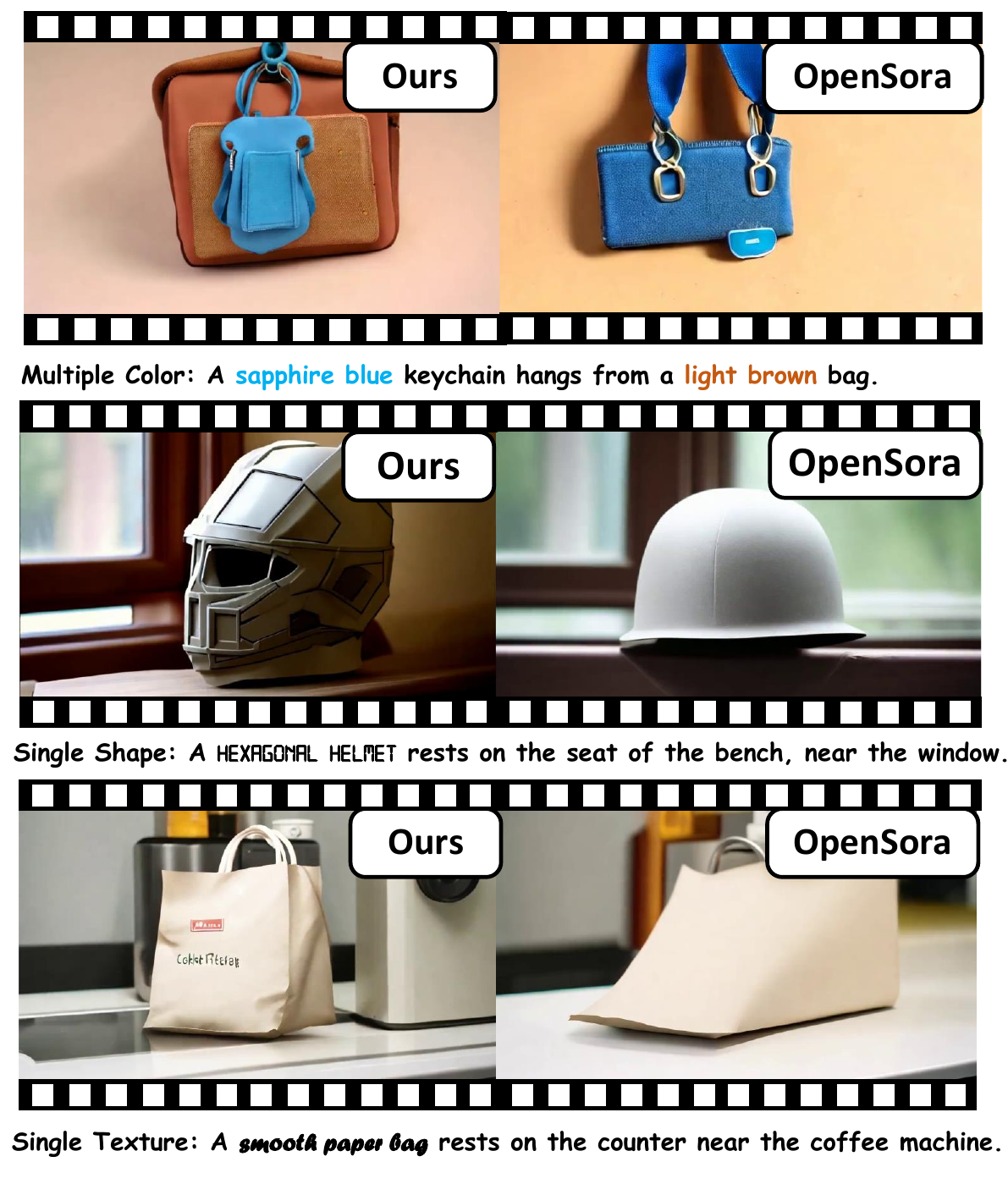}
	\vspace{-4mm}
	\caption{
		Visual comparison of~\oursName~and Open-Sora on \textit{instance-level attributes}.~\oursName~excels in precise instance detail fidelity and instruction-following capabilities, even with complex multi-instance and multi-attribute scenarios.
	}
	\label{fig:detail}
	\vspace{-2mm}
\end{figure}

\subsection{Text-to-video generation}
\paragraph{Qualitative evaluation.} Figures~\ref{fig:action} and~\ref{fig:detail} provide visual comparisons of the T2V generation results.  
It can be observed that the infusion of our~\oursData~dataset into Open-Sora \cite{opensora} serves to further enhance its video synthesis capabilities across four fundamental dimensions (Action, Color, Shape, and Texture). These four different aspects correspond to information in our instance-aware
structured captions such as ``Actions and Motion", ``Appearance", etc.
For instance, our model accurately generates the ``sprints" action of the lion in Figure~\ref{fig:action}, as opposed to Open-Sora~\cite{opensora}.
In Figure~\ref{fig:detail}, benefiting from our instance-aware caption, our model generates the accurate ``light brown bag" instance described in caption, where Open-Sora~\cite{opensora} completely loses this instance. These results indicate that our~\oursData~can provide accurate and instance-level guiding information for video generation models.

\begin{figure*}[!ht]
    \centering
    \includegraphics[width=0.99\linewidth]{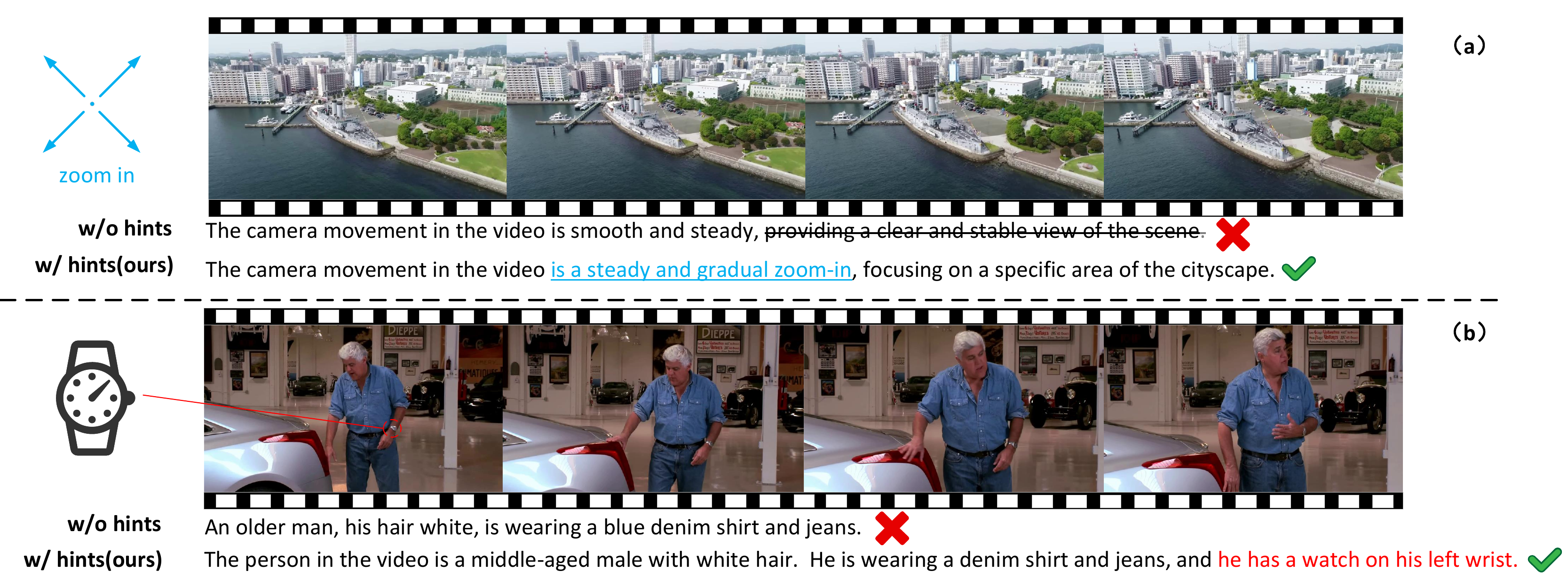}
    \vspace{-4mm}
    \caption{(a) Ablation study on the effect of camera movement hints on the accuracy of MLLM labeling. (b) Impact of human-designed class hints on the details of instance labeling.}
    \label{fig:camera_human}
    \vspace{-2mm}
\end{figure*}

\begin{figure*}[!ht]
    \centering
    \includegraphics[width=1\linewidth]{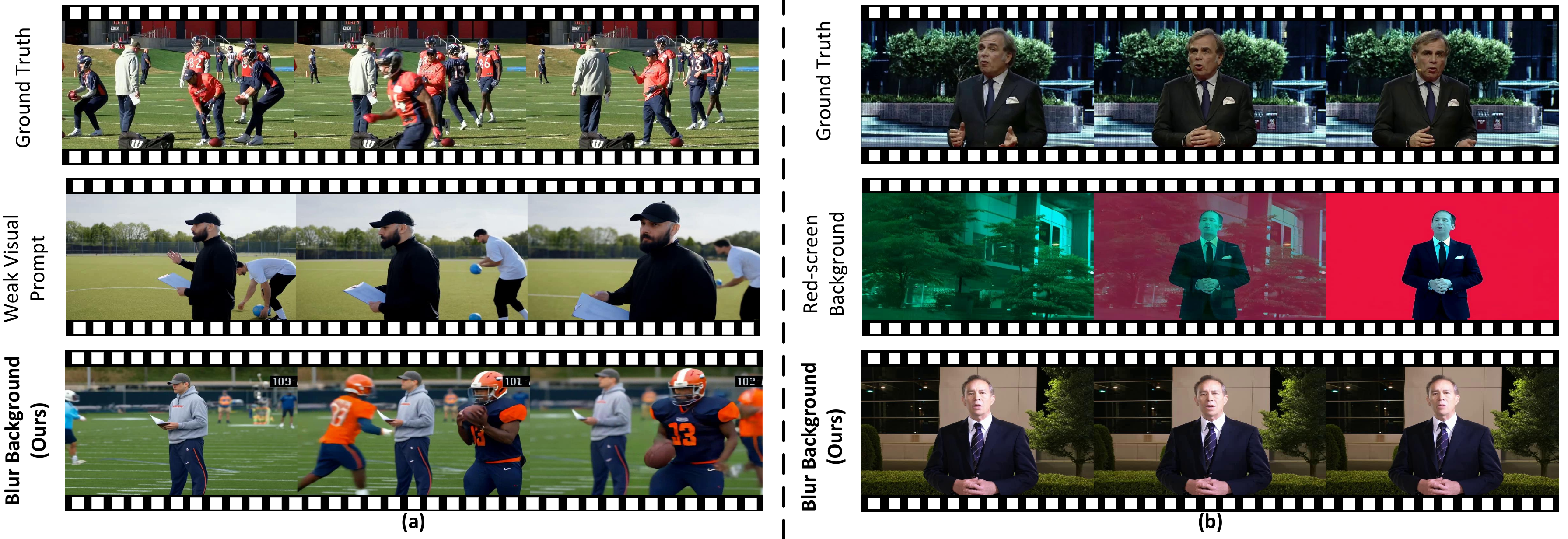}
    \vspace{-6mm}
    \caption{(a) Comparison against the weak visual prompt for reconstruction-via-caption visualization on multi-instance targets. (b) Comparison against color screen backgrounds (red), which may negatively affect MLLM labeling performance.}
    \label{fig:background}
    \vspace{-5mm}
\end{figure*}

\vspace{-4mm}
\paragraph{Quantitative evaluation.}
%
We conduct the quantitative evaluation for~\oursName~using the proposed Inseval metrics in Table~\ref{tab:inseval}. We can draw the following conclusions:
%
%
1) Fine-tuning with~\oursData~consistently improves all metrics over the base model Open-Sora, demonstrating the effectiveness of~\oursName.  
In particular, our Detail score ranks first, justifying the capacity of~\oursName~to capture complex instance detail in video.
2) Compared to other video captioning models finetuned based on Open-Sora,~\oursName~shows clear advantages in video generation tasks.
3) Compared to larger models like CogVideoX or Pyramid-Flow, our approach achieves a higher average metric than Pyramid-Flow, and performs comparably to CogVideoX in several specific metrics like `Single-Color/Shape/Detail' and `Multiple-Action', but with much fewer parameters. 


\subsection{Ablation Study}

\paragraph{Effects of human design class hints and camera movement hints.}
We discuss the impact of incorporating human-designed class hints and camera movement hints on annotation outcomes and provide relevant caption visualizations in Figure~\ref{fig:camera_human}. 
These annotations aid MLLMs in focusing more precisely on key elements, resulting in richer and more accurate annotations.

\vspace{-4.5mm}
\paragraph{Ablations on different video visual prompts.}
Comparative results for various video visual prompt methods used in caption generation are shown in Figure~\ref{fig:background}.
As illustrated in (a), weak visual prompts derived from static image techniques, such as red circles, bounding boxes, or selective grayscale manipulation of non-target areas~\cite{YAO202430,Shtedritski_2023_ICCV}, limit MLLMs in distinguishing and describe specific targets in multi-instance scenes, leading to attribute blending and vague annotations across instances. 
In contrast, our method excels in instance-specific feature extraction, accurately differentiating figures like the coach and players.
Figure~\ref{fig:background}~(b) illustrates strong visual prompts that involve complete occlusion of non-target regions to eliminate MLLM interactions with irrelevant instances. 
Conventional methods use primary color screens, but this often misguides MLLMs, causing them to incorporate incorrect context in captions. 
Our designed blur background masking approach, however, preserves visual consistency with natural scenes, enabling MLLMs to generate accurate and contextually relevant annotations with minimal prompting guidance.

\section{Conclusions and Limitations}

In this paper, we introduce~\oursName, the first instance-aware structured caption method for text-to-video generation.
We design an Auxiliary Models Cluster (AMC) to convert global video into instances, enhancing instance fidelity. 
we also propose an improved CoT pipeline with MLLMs to refine dense prompts into structured phrases, achieving concise yet precise instance descriptions compared to the previous video caption models.
Additionally, based on~\oursName, we curated~\oursData~dataset for training and~\oursEnhancer~during inference, significantly enhancing T2V models' generation capabilities on instance details and actions. 

\paragraph{Limitations.} 
Since the precision of~\oursName~partly depends on object detection methods, requiring fine-tuning of the detection model for domain-specific instances, and its benefits decrease in instance-free scenes.
Furthermore, the scale of~\oursData~limits its use as a large-scale pre-training dataset. 
Moving forward, we plan to apply~\oursName~to a larger video dataset and train more powerful T2V models to amplify its impact.

{
    \small
    \bibliographystyle{ieeenat_fullname}
    \bibliography{main}
}

\clearpage
\setcounter{page}{1}
\setcounter{table}{0} 
\setcounter{figure}{0}
\setcounter{section}{0}

\renewcommand{\thetable}{S\arabic{table}}
\renewcommand{\thefigure}{S\arabic{figure}}

\maketitlesupplementary
In this supplementary material, we present comprehensive details and analyses across the following sections:

\begin{itemize}
\item \textbf{Section}~\ref{supp:lex} elucidates our methodology for constructing positive/negative lexical databases, accompanied by their details.

\item \textbf{Section}~\ref{supp:ch} provides an extensive compilation of human-designed class hints, demonstrating their diverse applications.

\item \textbf{Section}~\ref{supp:fg3} delineates the improved Chain-of-Thought prompting strategies employed in Figure~\ref{fig:mllm}, with particular emphasis on their methodological improvements.

\item \textbf{Section}~\ref{supp:enhancer} explicates the architectural framework of~\oursEnhancer, supplemented with exemplary prompts utilized in our Large Language Model implementations.

\item \textbf{Section}~\ref{supp:video_recon_eval} elaborates a detailed discussion of the principles behind our metric design for video reconstruction, including mathematical formulations and empirical validations.

\item \textbf{Section}~\ref{supp:inseval} demonstrates the prompts used by Inseval in both the inference and evaluation stages.

\item \textbf{Section}~\ref{supp:analysis} presents a evaluation of our methodology across both commercial and open-source models, including experimental results and analytical findings.
\end{itemize}

\section{Positive/Negative Lexicon}
\label{supp:lex}
\begin{figure}[ht]
    \centering
    \includegraphics[width=1\linewidth]{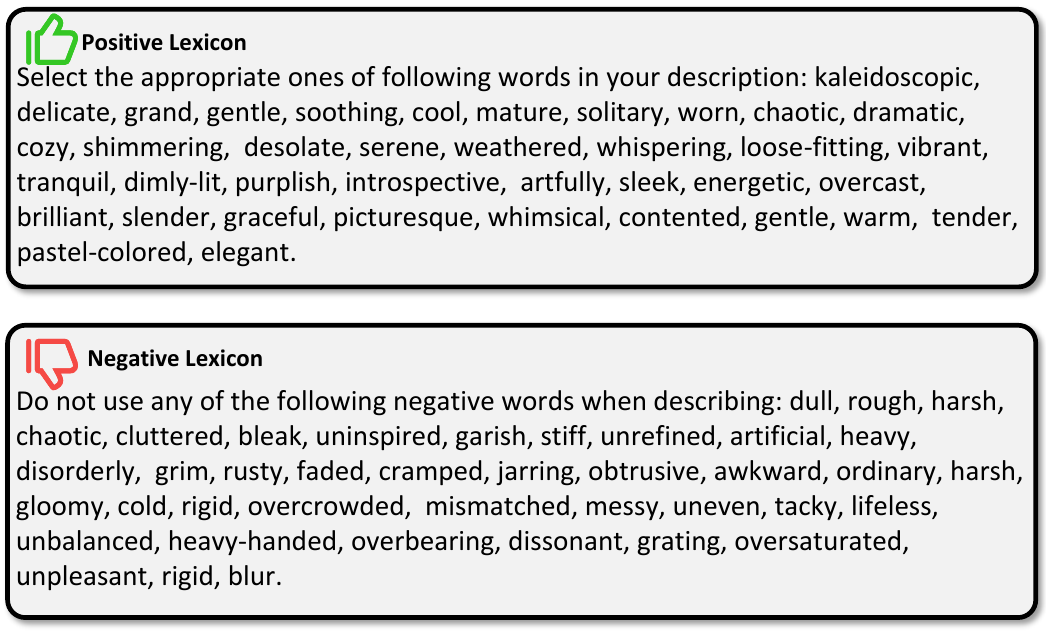}
    \caption{The detail of Positive/Negative Lexicon}
    \label{fig:np}
\end{figure}

To enhance the aesthetic quality of generated videos, we carefully collected prompts from various open-source model galleries, extracting adjectives to build a \textit{Positive Lexicon}.
Conversely, we manually constructed a \textit{Negative Lexicon}, which was further enriched using the powerful LLM, GPT-4o. 
Both lexicons were refined through meticulous manual screening. 
The detailed contents of the Positive/Negative Lexicons are shown in Figure~\ref{fig:np}.

\section{Human-designed Class Hints}
\label{supp:ch}
For the Human-designed Class Hints, we carefully crafted additional prompts for over \textit{eighty} categories, each specifically tailored to its specific characteristics. Below, we present twenty of these categories. 
The full JSON-formatted hints for all classes, ready for direct use, will be provided in the code we plan to release later.
\begin{itemize}
    \item  \textbf{Person}: ``Please focus primarily on the person's facial expressions, attire, age, gender, and race in the video and give description in detail. Please mention if there are any necklaces, watches, hat or other decoration; otherwise, there’s no need to bring them up.''
    \item  \textbf{Bicycle}: ``Please describe the bicycle in terms of color, type, size, condition, and any distinctive marks or decorations. Include details such as the presence of baskets, reflectors, or any branding.''
    \item  \textbf{Car}: ``Please describe the car by its color, make, model, condition, license plate (if visible), and any distinguishing features such as stickers, dents, or modifications.''
    \item  \textbf{Airplane}: ``Please describe the airplane by its type (commercial, private, etc.), airline brand, color scheme, size, and any visible markings such as logos or tail numbers.''
    \item  \textbf{Bus}: ``Please describe the bus by its color, type (public, school, etc.), condition, any branding or advertising on its surface, and the route number or destination if visible.''
    \item  \textbf{Train}: ``Please describe the train by its type (freight, passenger, high-speed, etc.), color, length, condition, and any visible logos or car numbers.''
    \item  \textbf{Truck}: ``Please describe the truck by its type (pickup, semi, etc.), color, make, model, any visible logos or branding, and details such as cargo or modifications.''
    \item  \textbf{Boat}: ``Please describe the boat by its type (sailboat, motorboat, yacht, etc.), size, color, condition, and any identifying features like registration numbers or flags.''
    \item  \textbf{Traffic Light}: ``Please mention the current state of the traffic light (red, yellow, green), its location, and any additional details like the presence of pedestrian signals.''
    \item  \textbf{Fire Hydrant}: ``Please describe the fire hydrant by its color, condition, and any notable features such as signs, markings, or proximity to other objects.''
    \item  \textbf{Stop Sign}: ``Please describe the stop sign's condition, location, and any visible obstructions or markings on it.''
    \item  \textbf{Parking Meter}: ``Please describe the parking meter by its condition, type (modern, traditional), and any visible information like pricing or operational status.''
    \item  \textbf{Bench}: ``Please describe the bench by its material, color, condition, and any distinctive features such as inscriptions, decorations, or nearby objects.''
    \item  \textbf{Bird}: ``Please describe the bird by its species (if identifiable), color, size, behavior, and any unique markings or features.''
    \item  \textbf{Cat}: ``Please describe the cat by its color, breed (if identifiable), size, behavior, and any distinguishing features such as collars or patterns.''
    \item  \textbf{Dog}: ``Please describe the dog by its breed (if identifiable), color, size, behavior, and any accessories such as collars or leashes.''
    \item  \textbf{Horse}: ``Please describe the horse by its color, breed (if identifiable), size, behavior, and any accessories such as saddles or reins.''
    \item  \textbf{Sheep}: ``Please describe the sheep by its color, size, behavior, and any distinguishing features such as markings or tags.''
    \item  \textbf{Cow}: ``Please describe the cow by its color, breed (if identifiable), size, behavior, and any distinguishing features such as tags or markings.''
    \item  \textbf{Elephant}: ``Please describe the elephant by its size, tusk length, condition, and any unique features such as markings or behavior.''
\end{itemize}

\begin{figure*}[!ht]
    \centering
    \includegraphics[width=1\linewidth]{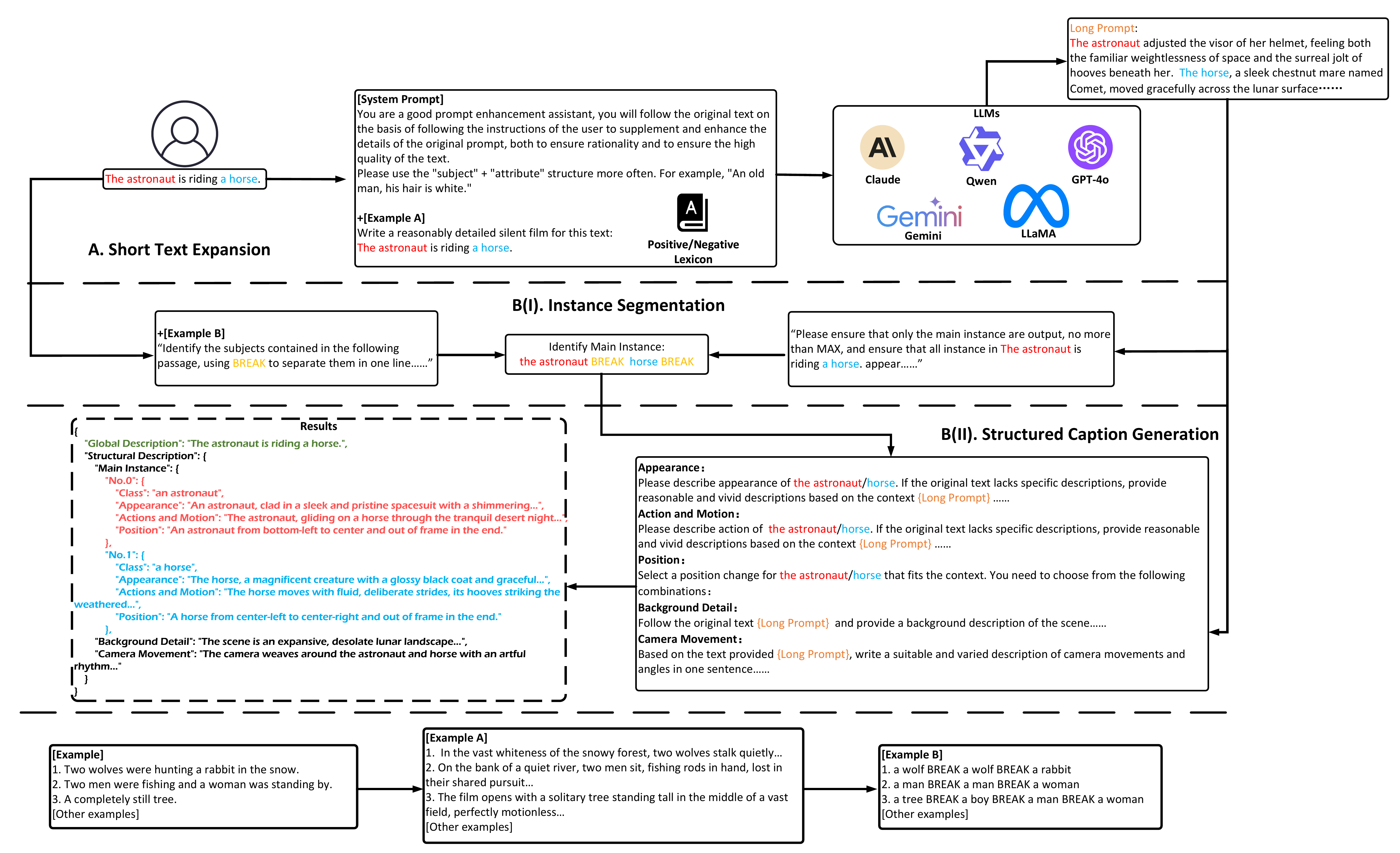}
    \caption{Detailed overview of the~\oursEnhancer~pipeline. Example No.1 as shown in Figure~\ref{fig:example}.}
    \label{fig:enhancer}
\end{figure*}

\section{Prompt Design of Figure~\ref{fig:mllm}}
\label{supp:fg3}
\begin{figure*}[ht]
    \centering
    \includegraphics[width=1\linewidth]{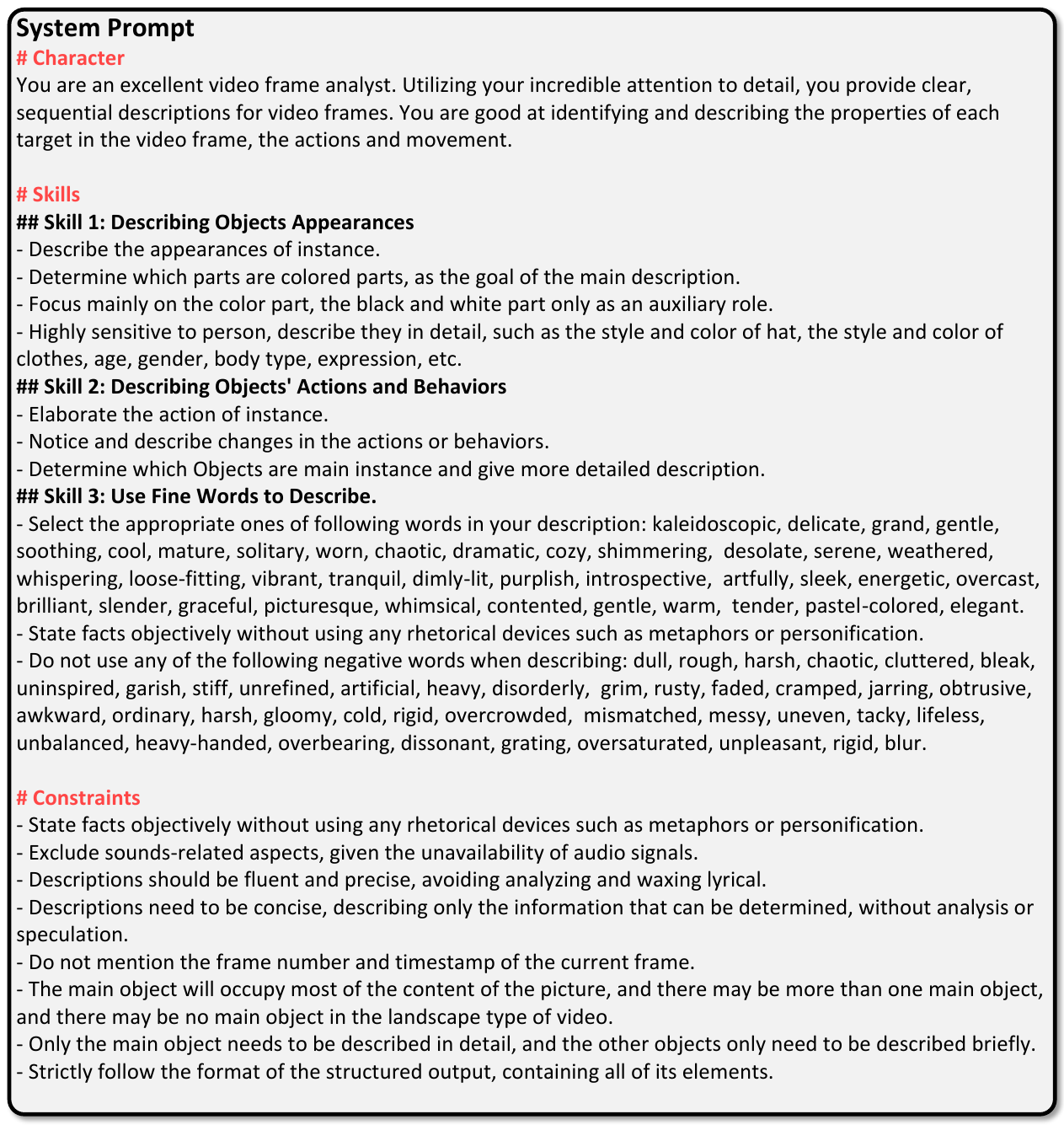}
    \caption{System prompt of~\oursName.}
    \label{fig:system}
\end{figure*}

\begin{figure}[!t]
    \centering
    \includegraphics[width=1\linewidth]{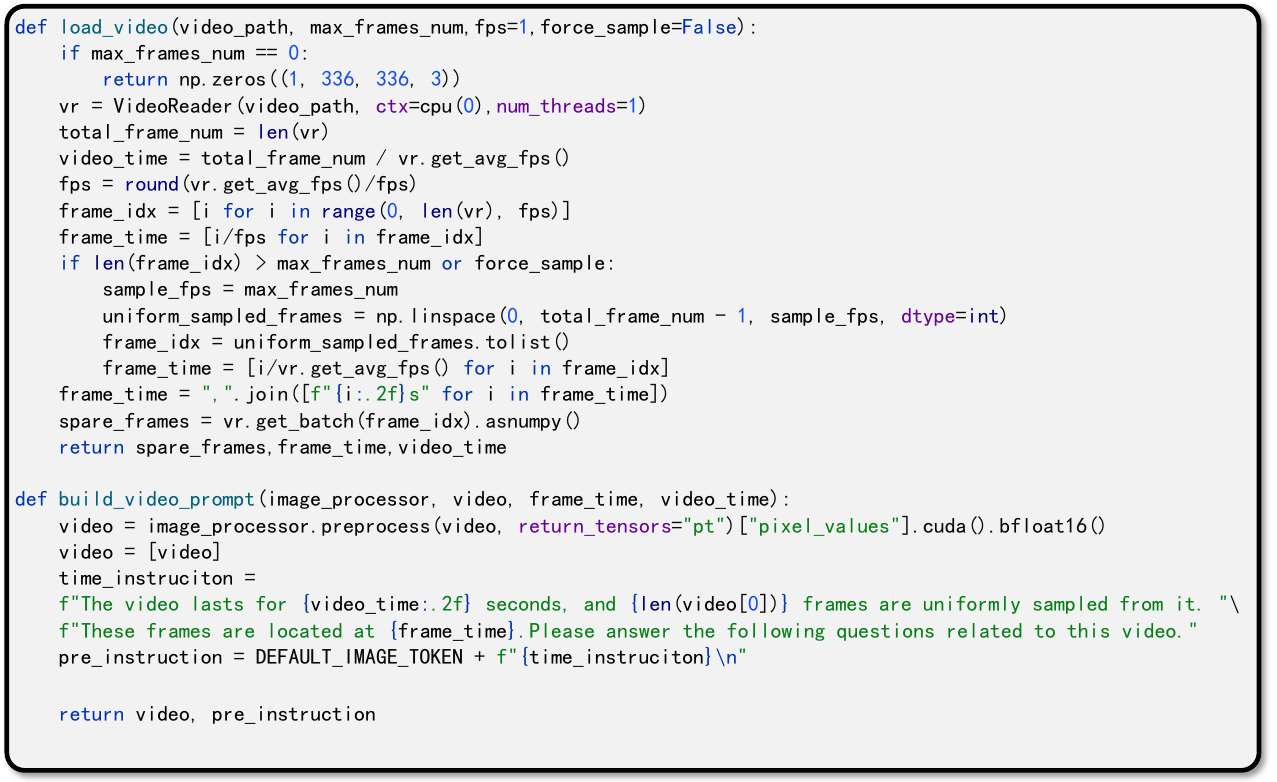}
    \caption{Code of getting video temporal metadata.}
    \label{fig:code}
\end{figure}

\paragraph{System prompt.}
Referring to ShareGPT4Video~\cite{chen2024sharegpt4video}, we divided the System prompt into three parts. Through extensive tests on challenging samples, including multi-instance, complex scenes, and high-intensity motion, we finalized the system prompt shown in Figure~\ref{fig:system}. 
Additionally, temporal metadata extracted using the code provided in Figure~\ref{fig:code}.

\begin{figure}[ht]
    \centering
    \includegraphics[width=1\linewidth]{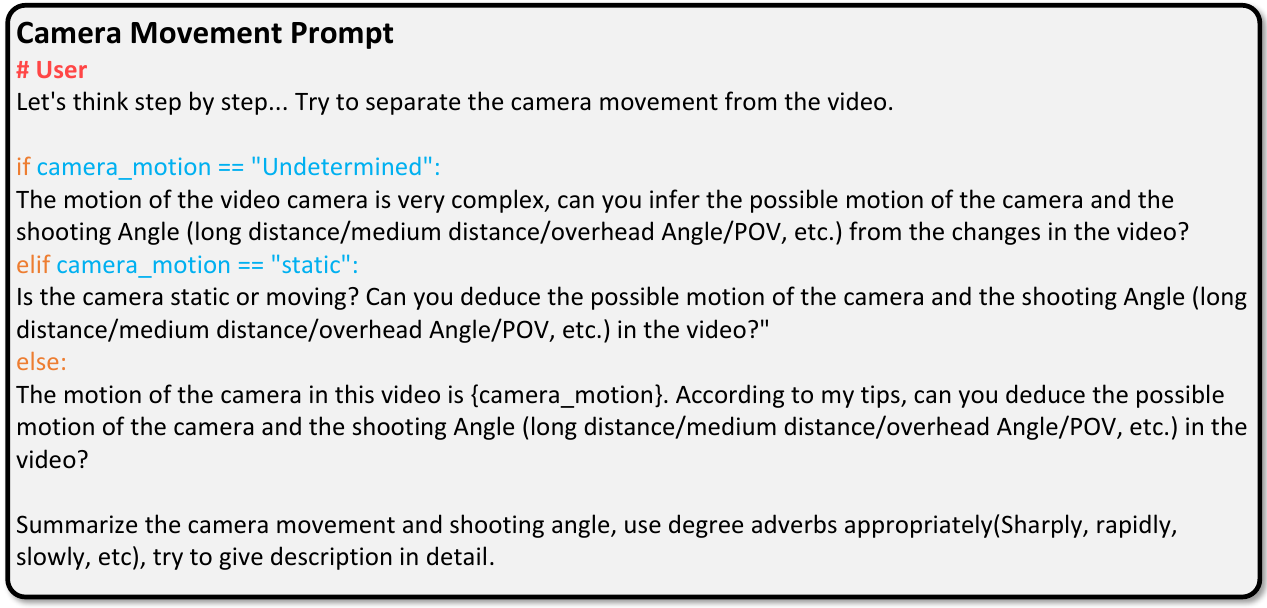}
    \caption{Prompt of camera movement.}
    \label{fig:camera}
\end{figure}

\paragraph{Prompts of global description, background detail and camera movement.}  
The global description is derived from a single prompt: \textit{``Please describe this video in one sentence, no more than 20 words.''}. 
To illustrate the acquisition of camera motion and background details, we provide an example of implementing camera hints with movement cues in Figure~\ref{fig:camera}.
A similar approach is used for extracting background details included in our code released later.

\begin{figure}[!ht]
    \centering
    \includegraphics[width=1\linewidth]{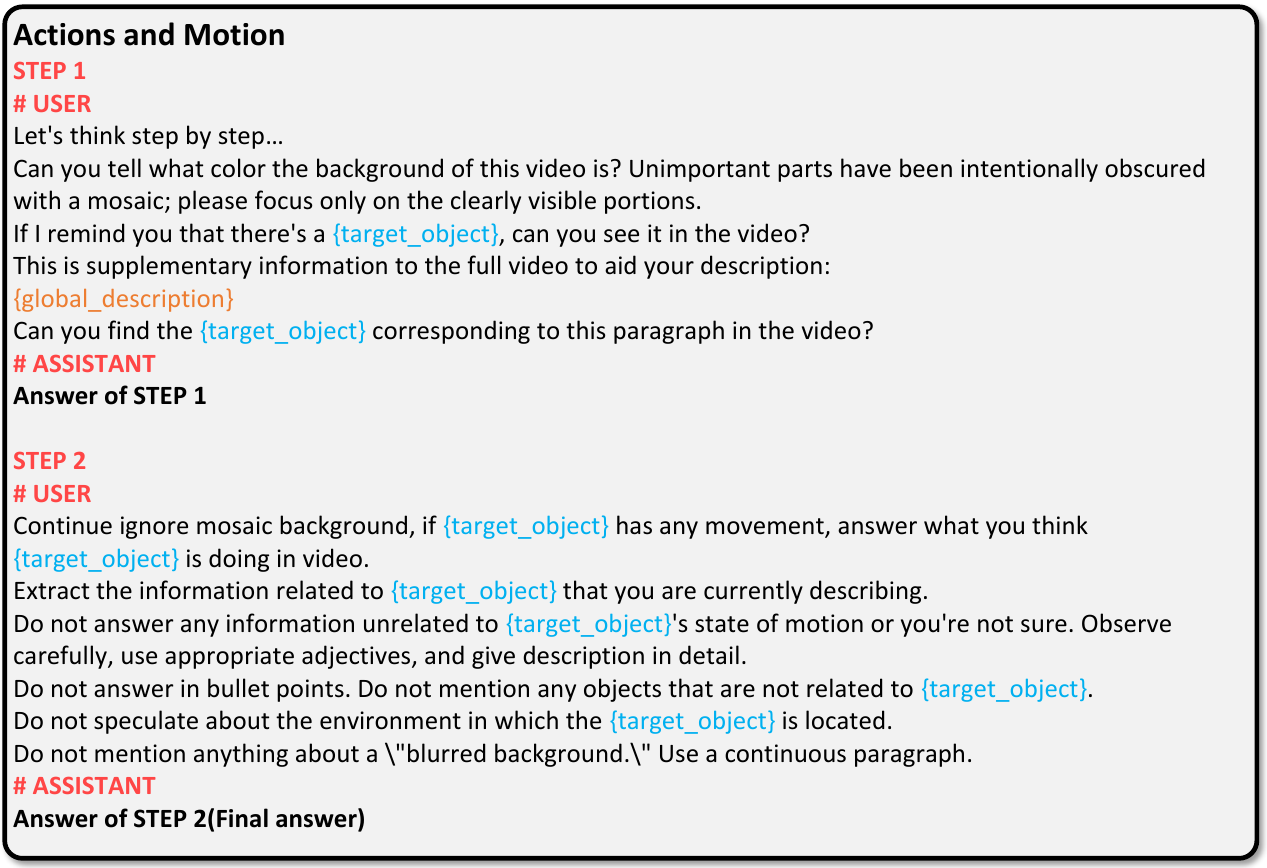}
    \caption{Prompt of actions and motion.}
    \label{fig:action_prompt}
\end{figure}

\paragraph{Prompts of structured caption.}
In the structured caption section, we we use Actions and Motion as examples, with the CoT prompt shown in Figure~\ref{fig:action_prompt}. 
The acquisition of Appearance and the injection of Human-designed class hints follow a similar approach.

\section{Design of InstanceEnhancer}
\label{supp:enhancer}
\begin{figure}[!ht]
    \centering
    \includegraphics[width=1\linewidth]{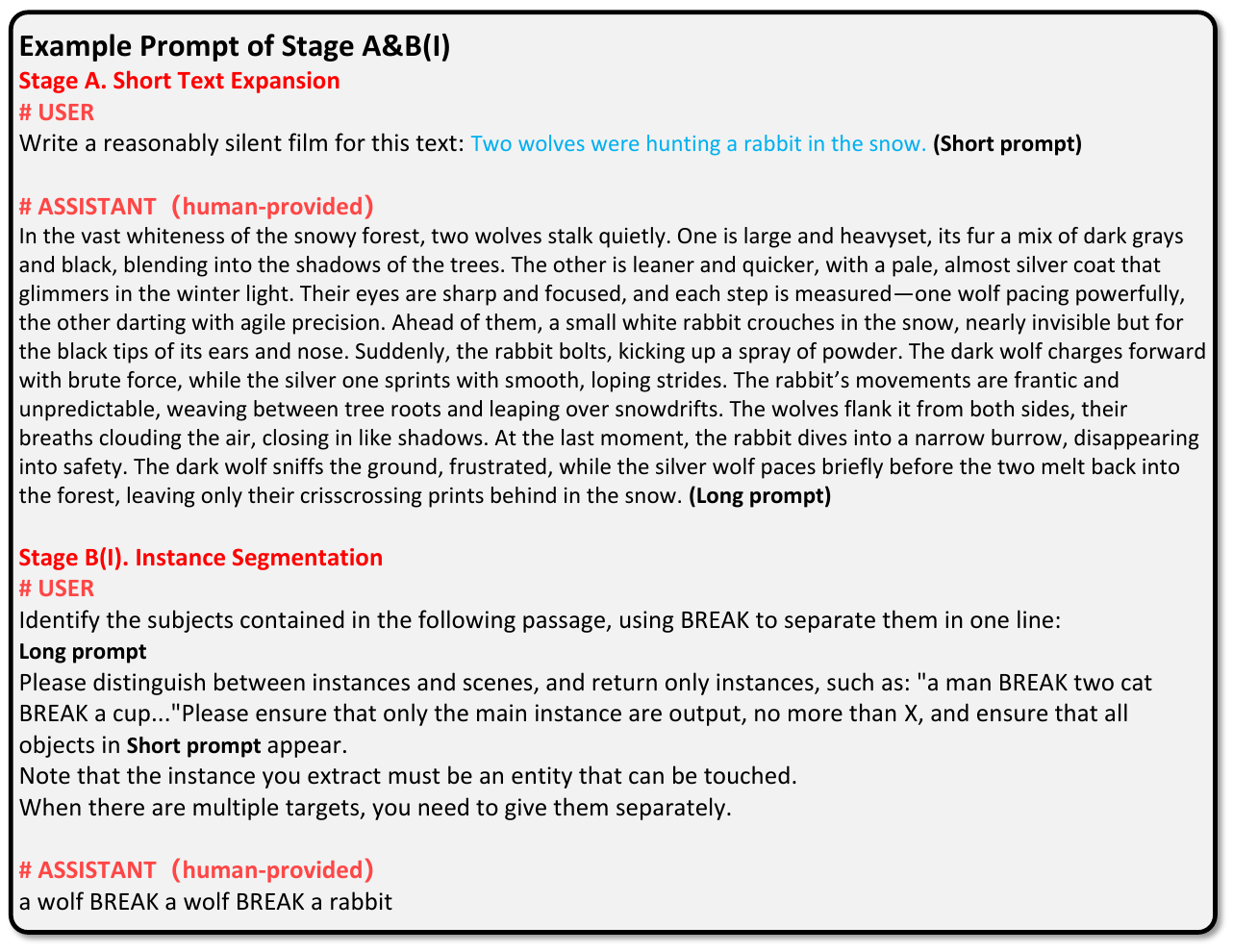}
    \caption{Designed example for LLMs.}
    \label{fig:example}
\end{figure}

In~\oursEnhancer, prompt alignment during inference is achieved through a two-stage process (Figure~\ref{fig:enhancer}). 
To provide more precise instructions to LLMs, we meticulously designed multiple examples as part of the CoT, which are fed into the LLMs. An example of this is shown in Figure~\ref{fig:example}.

\section{Evaluation metrics for video reconstruction}
\label{supp:video_recon_eval}
\paragraph{3DVAE score ($\text{3DVAE}_{score}$).}
The LIPIPS score \cite{zhang2018unreasonableeffectivenessdeepfeatures} which is widely used to evaluate image reconstruction quality, measures perceptual distance between ground truth (GT) and reconstructed images. 
We extent this concept for video data by using 3DVAE \cite{yang2024cogvideox} to extract latent-space video representations from both GT videos and their caption-reconstructed versions. $\text{3DVAE}_{score}$ computes the distance between latent representations across spatial and temporal dimensions:

\begin{equation}
d(\mathbf{z}_{\text{GT}}, \mathbf{z}_{\text{rec}}) = \sum_{l} \sum_{t} \sum_{h, w} \left\| w_l \odot \left( \mathbf{z}_{\text{GT}, hwt}^l - \mathbf{z}_{\text{rec}, hwt}^l \right) \right\|_2^2
\end{equation}

where \( \mathbf{z}_{\text{GT}, hwt}^l \) and \( \mathbf{z}_{\text{rec}, hwt}^l \) represent the latent representations at layer \( l \), spatial location \( (h, w) \), and temporal frame \( t \), with \( w_l \) as the layer-specific weight matrix. We set \((h, w, t) = (224, 224, 8)\) for evaluation.

To ensure consistency, we use the same video generation model across all captioning methods. Following LIPIPS methodology, we validate the 3DVAE score by comparing GT videos against various distorted versions. As shown in~\cref{tab:3dvae}, the results demonstrate that our score effectively captures perceptual similarities between GT and reconstructed videos.

\begin{table}[!ht]
\centering
\resizebox{.95\linewidth}{!}{%
\begin{tabular}{l|c|c}
\toprule
\textbf{Distortion type} & \textbf{3DVAE score↓} & \textbf{Setting} \\
\midrule
Blurring & 7.71 & GaussianBlur(kernel=(5, 5), sigma=0) \\
Compression artifacts & 11.19 & JPEG compression (quality 5-30) \\
Corruptions & 39.80 & Random pixel masking (binary mask) \\
Random noise & 49.70 & Gaussian noise (mean=0, stddev=25) \\
Brightness distortion & 63.25 & Scaling (factor 0.5-1.5) \\
Spatial shifts & 78.94 & Random affine shifts (±10 pixels) \\
\midrule
T2V models Avg. & 134 $\sim$ 145 & - \\
Broken video & 149.50 & - \\
\bottomrule
\end{tabular}%
}
\caption{3DVAE scores for various distortions and video models, showcasing its effectiveness in capturing perceptual similarities and reconstruction accuracy. The setting column provides details of the experimental setup for each distortion type.}
\label{tab:3dvae}
\end{table}

\begin{figure*}[!ht]
    \centering
    \includegraphics[width=1\linewidth]{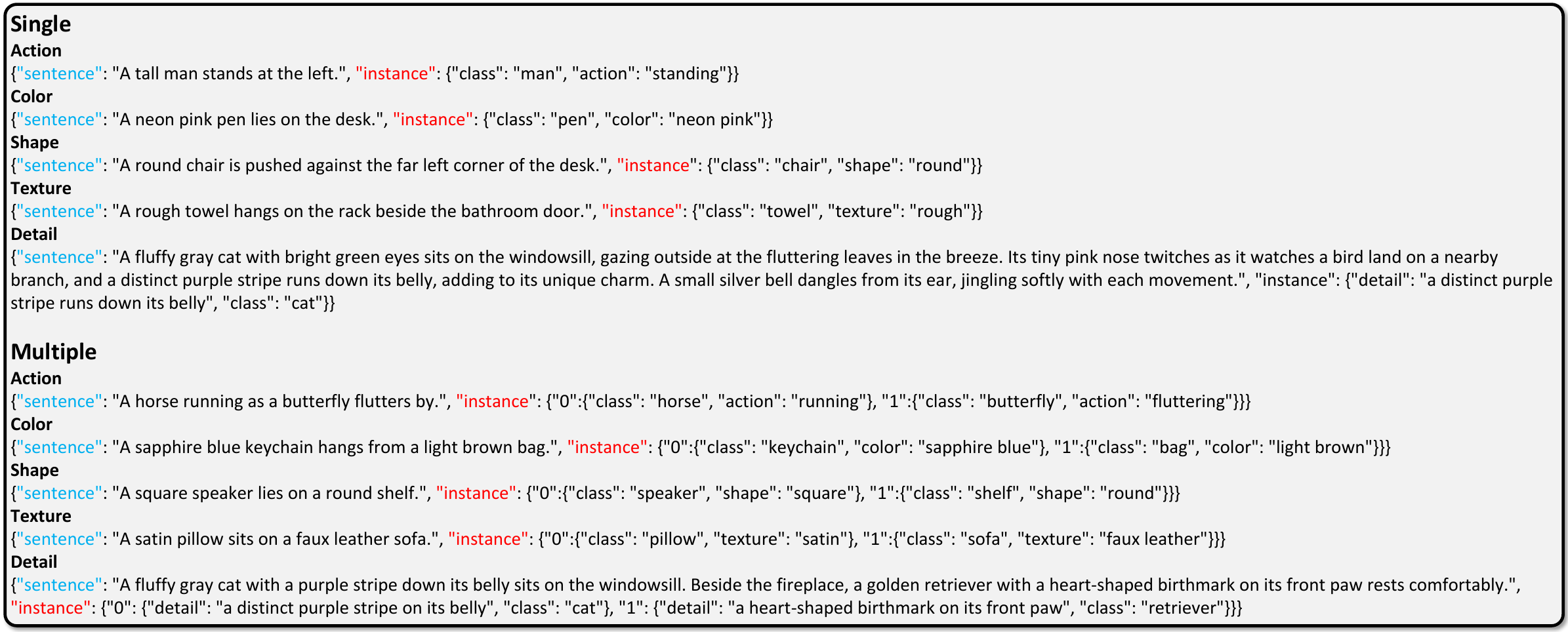}
    \caption{Inference examples of Inseval.}
    \label{fig:inseval_infer}
\end{figure*}

\begin{figure*}[!ht]
    \centering
    \includegraphics[width=.9\linewidth]{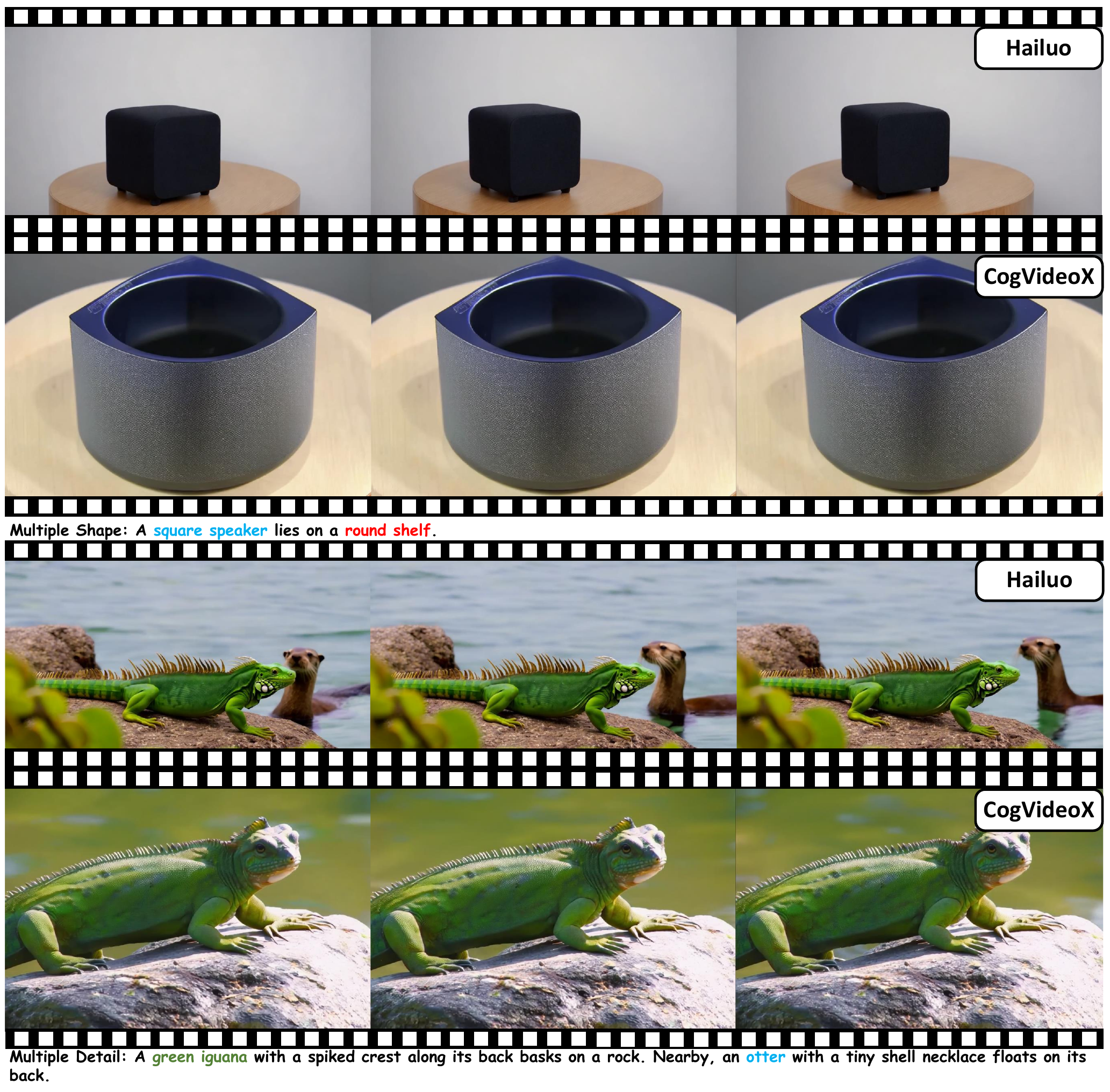}
    \caption{Visualization comparing open-source models and commercial models on prompts with poorer performance.}
    \label{fig:compare_multi}
\end{figure*}

\begin{table*}[!t]
\centering
\begin{tabular}{p{0.5cm} p{7cm} | p{0.5cm} p{7cm}}
\toprule
\multicolumn{2}{c|}{\textbf{Instance Detail}} & \multicolumn{2}{c}{\textbf{Hallucination Scores}} \\
\midrule
\textbf{1} & Descriptions are extremely vague, imprecise, or largely inaccurate. Almost no specific details from the video are captured correctly. & 
\textbf{1} & Severe hallucination - Describes many nonexistent details, significantly misrepresents what is shown, or introduces extensive irrelevant content with many unrelated topics or external information. \\
\midrule
\textbf{2} & Descriptions have major inaccuracies or omit many important details. Only a few basic elements are described correctly. & 
\textbf{2} & Frequent hallucination - Multiple instances of fabricated or misrepresented details and significant extra content introducing information beyond the video scope. \\
\midrule
\textbf{3} & Descriptions are moderately accurate but lack precision in some areas. Core details are present but some secondary details are missing or incorrect. & 
\textbf{3} & Occasional hallucination - A few minor instances of fabricated details, misrepresentations, or the addition of extra content not covered in the video. \\
\midrule
\textbf{4} & Descriptions are largely accurate and detailed. Most key elements and nuances from the video are captured correctly, with only minor omissions or imprecisions. & 
\textbf{4} & Minimal hallucination - One or two very minor discrepancies or limited introduction of external information. \\
\midrule
\textbf{5} & Descriptions are highly precise and comprehensive. All important details from the video are captured accurately, including subtle elements and specific examples. & 
\textbf{5} & No hallucination - All described details accurately reflect what is shown in the video, with no external content added. \\
\bottomrule
\end{tabular}
\caption{This table outlines scoring criteria for Instance Detail and Hallucination Scores, integrating intrinsic and extrinsic hallucinations into a unified framework for evaluation.}
\label{tab:id_hs}
\end{table*}

\paragraph{CLIP score sentence by sentence ($\text{CLIP}_{SenbySen}$).}
While CLIP~\cite{radford2021learningtransferablevisualmodels} is widely used for text-video similarity computation~\cite{liu2023evalcrafter, huang2023vbench}, its 77-token limit restricts processing of long texts. To overcome this, we propose CLIP score sentence by sentence (SenbySen), which segments texts into individual sentences and computes CLIP similarity between each sentence and video frame.

Let \( S = \{s_1, s_2, \ldots, s_n\} \) be the sentences from input text and \( V = \{v_1, v_2, \ldots, v_t\} \) be the video frames. For a sentence \( s_i \) and frame \( v_j \), we denote their CLIP similarity as \( \text{CLIP}(s_i, v_j) \). The comprehensive score is computed as:

\begin{equation}
\text{OverallScore} = \frac{1}{n} \sum_{i=1}^{n} \left( \frac{1}{t} \sum_{j=1}^{t} \text{CLIP}(s_i, v_j) \right)
\end{equation}

This approach not only addresses the token limitation but also enhances assessment quality by naturally assigning lower weights to non-specific textual descriptions.

\paragraph{Human evaluation.}
Automated machine-based scoring systems, while offering enhanced objectivity and efficiency, often fail to align with human preferences or fully grasp the nuances of context and meaning in a given task.
To ensure a comprehensive and balanced evaluation, we adopted a human-based assessment framework. This evaluation is carried out across several key dimensions, including:
    1) Instance Detail (\textbf{ID}): Evaluate whether the text provides accurate descriptions of the details of the examples in the video.
    2) Intrinsic Hallucination: Evaluate whether the text hallucinates descriptions of things present in the video.
    3) Extrinsic Hallucination: Evaluate whether the text introduces content that is not present in the video.
    For convenience, the latter two have been combined into a single metric called the Hallucination Scores (\textbf{HS})~\cite{Ji_2023}. 
    The specific guidelines and scoring criteria for each metric refers to Table~\ref{tab:id_hs}.

\section{Inseval}
\label{supp:inseval}
\paragraph{Inference prompts of Inseval.}
In implementing Inseval, we designed multiple prompts to test each dimension, as illustrated in Figure~\ref{fig:inseval_infer}. 
To further evaluate the model's generative capabilities and instruction-following accuracy, we deliberately included some ``counter-intuitive" shapes in the prompt design.

\begin{figure}[!ht]
    \centering
    \includegraphics[width=1\linewidth]{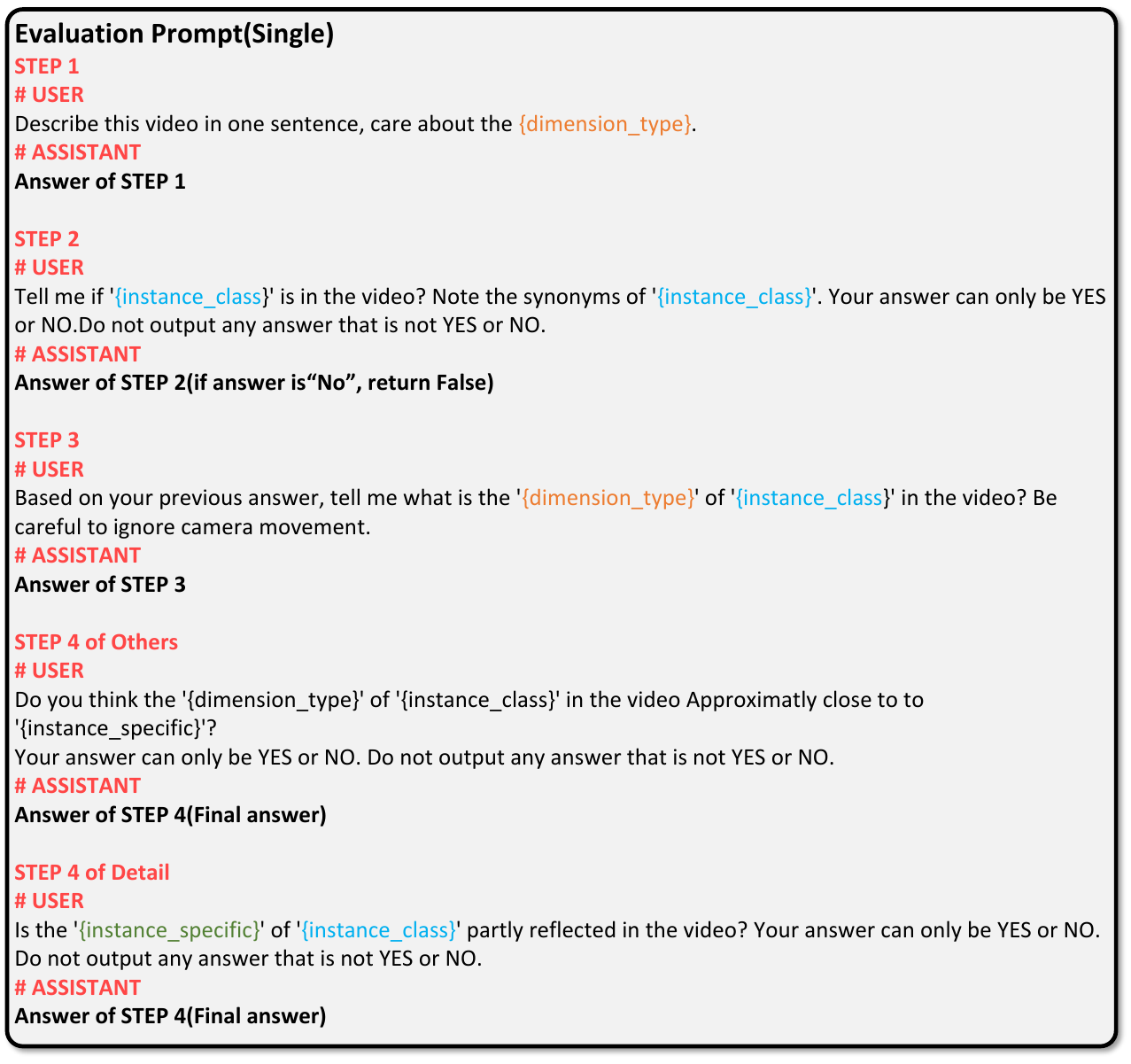}
    \caption{Evaluation prompts of Inseval.}
    \label{fig:inseval_eval}
\end{figure}

\paragraph{Evaluation prompts of Inseval.}
For the evaluation, we used a general CoT Q-A pair format (with a slightly different design for the `Detail' dimension, shown in Figure~\ref{fig:inseval_eval} to assess whether the MLLMs successfully matched the generated videos to the corresponding dimensions, as outlined in the specific code.
In single-object scenarios, the success rate is calculated as the percentage of correctly matched prompts. 
In multi-object scenarios, the generation is deemed successful only if all targets meet the requirements. 
For reproducibility, fixed random seeds are used during generation and evaluation.


In Table~\ref{tab:inseval}, the `Shape' and `Detail' dimensions under Multiple category are omitted due to consistently very poor performance across all tested models. 
Even CogVideoX-5B, the overall best performer, struggles with multi-object tasks in these dimensions, as shown in Figure~\ref{fig:compare_multi}. 
Two primary error types are observed in Multiple Shape tasks: attribute confusion (\textbf{Top} case) and failure to follow multiple target instructions (\textbf{Bottom} case), where targets are either missing or rendered incorrectly. Commercial models demonstrate relatively better performance, which we further analyze in Section~\ref{supp:analysis}.

\section{Analysis on Commercial Products vs. Open-source Models}
\label{supp:analysis}
\paragraph{Prompt processing analysis.}
Commercial T2V products excel at processing complex input prompts, effectively handling long-form text in structured formats while preserving semantic coherence. They can seamlessly interpret detailed scene descriptions, character interactions, and sequential events within a single prompt, producing coherent visual narratives, have shown surprising results in many situations.

\begin{figure*}[!ht]
    \centering
    \includegraphics[width=1\linewidth]{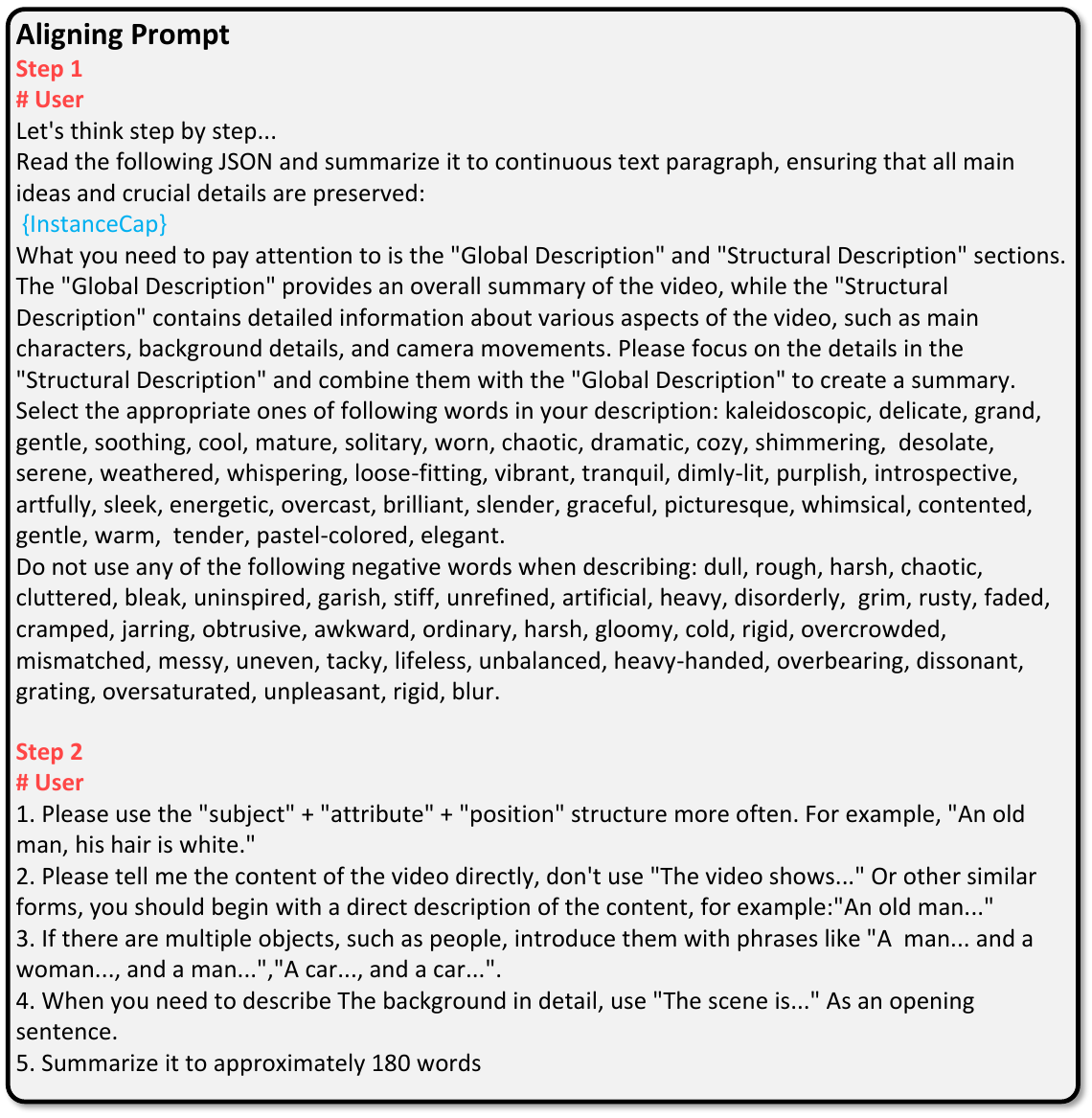}
    \caption{Aligning prompt used during alignment with the open source model.}
    \label{fig:align}
\end{figure*}

Open-source T2V models, however, are \textit{unable to directly process long-text structured prompts}, requiring an additional alignment step (Figure~\ref{fig:align}).
This preprocessing can lead to potential information loss and inconsistencies in the final output, restricting the ability to capture nuanced details from the original prompt.

\paragraph{Information retention capabilities.}
Different models exhibit notable differences in information retention (Figure~\ref{fig:compare_multi}). 
Commercial products (\textit{e.g.}, Hailuo AI) excel in maintaining fidelity between text and visual content, effectively preserving detailed instructions and translating multiple attributes into video sequences.
This strength is particularly apparent when our caption contains \textit{complex scenes} that demand temporal consistency and fine-grained details.


Open-source models face challenges in consistently representing instance information (Figure~\ref{fig:compare_multi}), exhibiting variability in detail preservation and limited capability with complex attribute combinations. 
These shortcomings are particularly evident when processing prompts with multiple interrelated instances or maintaining consistent visual characteristics across temporal sequences.

\end{document}